\documentclass[runningheads]{llncs}
\usepackage{graphicx}
\graphicspath{{pdf/}{jpeg/}}
\DeclareGraphicsExtensions{.pdf,.jpeg,.png}
\usepackage[caption=false,labelfont=sf]{subfig}
\begin{document}

\title{Design Pseudo Ground Truth with Motion Cue for Unsupervised Video Object Segmentation} 

\titlerunning{Design Pseudo Ground Truth with Motion Cue for Unsupervised VOS} 


\author{Ye Wang\inst{1} \and
Jongmoo Choi\inst{1} \and 
Yueru Chen\inst{1} \and
Qin Huang\inst{1} \and 
Siyang Li\inst{1} \and
Ming-Sui Lee\inst{2} \and 
C.-C. Jay Kuo\inst{1}}
%

\authorrunning{Y. Wang et al.} 



\institute{University of Southern California, Los Angeles CA 91754, USA \\
\email{\{wang316, jongmooc, yueruche, qinhuang, siyangl, cckuo\}@usc.edu }\\
\and
National Taiwan University, Taipei 10617, Taiwan\\
\email{mslee@csie.ntu.edu.tw}}

\maketitle

\begin{abstract}
One major technique debt in video object segmentation is to label the object masks for training instances. As a result, we propose to prepare inexpensive, yet high quality pseudo ground truth corrected with motion cue for video object segmentation training. Our method conducts semantic segmentation using instance segmentation networks and, then, selects the segmented object of interest as the pseudo ground truth based on the motion
information.  Afterwards, the pseudo ground truth is exploited to
finetune the pretrained objectness network to facilitate object
segmentation in the remaining frames of the video.  We show that the
pseudo ground truth could effectively improve the segmentation
performance. This straightforward unsupervised video object segmentation
method is more efficient than existing methods. Experimental results on DAVIS and FBMS show that the proposed method outperforms state-of-the-art unsupervised
segmentation methods on various benchmark datasets. And the category-agnostic pseudo ground truth has great potential to extend to multiple arbitrary object tracking.

\end{abstract}
\section{Introduction}\label{sec:intro}

Video object segmentation (VOS) is the task to segment foreground
objects from background across all frames in a video clip. The VOS
m{}ethods can be classified into two categories: semi-supervised and
unsupervised VOS methods. Semi-supervised VOS methods \cite{marki2016bilateral,tsai2016video,voigtlaender2017online,caelles2017one,perazzi2017learning} require the ground truth
segmentation mask in the first frame as the input and, then, segment the
annotated object in the remaining frames. Unsupervised VOS methods
\cite{koh2017primary,tokmakov2017learning,jain2017fusionseg,cheng2017segflow,li2018instance,li2018unsupervised} identify and segment the main object in
the video automatically. 

Recent image-based semantic and instance segmentation tasks
\cite{he2017mask,huang2017semantic,huang2018unsupervised} have
achieved great success due to the emergence of deep neural networks such
as the fully convolutional network (FCN) \cite{long2015fully}.  The
one-shot video object segmentation (OSVOS) method \cite{caelles2017one}
uses large classification datasets in pretraining and applies the
foreground/background segmentation information obtained from the first
frame to object segmentation in the remaining frames of the video clip.
It converts the image-based segmentation method to a semi-supervised
video-based segmentation method by processing each frame independently
without using the temporal information. 

However, since manual annotation is expensive, it is desired to develop
the more challenging unsupervised VOS solution. This is feasible due to
the following observation. Inspired by vision studies
\cite{port1995mind}, moving objects can attract infant and young
animals' attention who can group things properly without knowing what
kinds of objects they are. Furthermore, we tend to group moving objects
and separate them from background and other static objects. In other
words, semantic grouping is acquired after motion-based grouping in the
VOS task. 

\begin{figure}[t]
\captionsetup[subfigure]{}
\centering 
\subfloat{\includegraphics[width=0.24\linewidth]{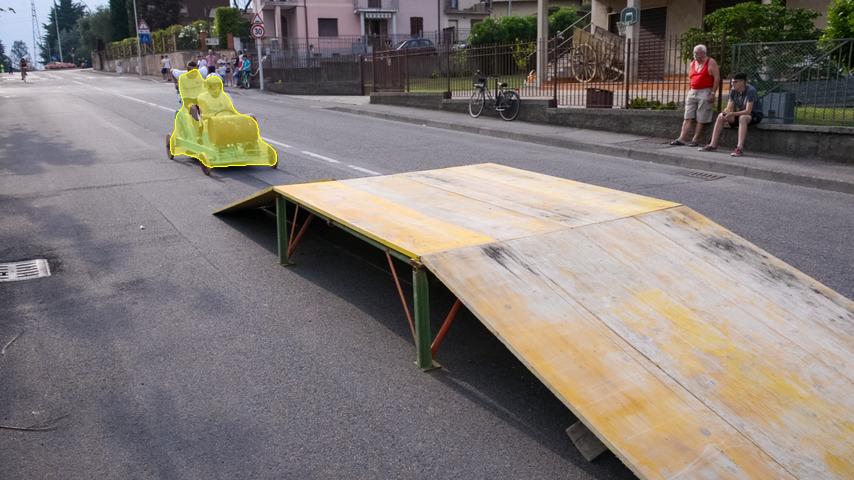}} \hfil 
\subfloat{\includegraphics[width=0.24\linewidth]{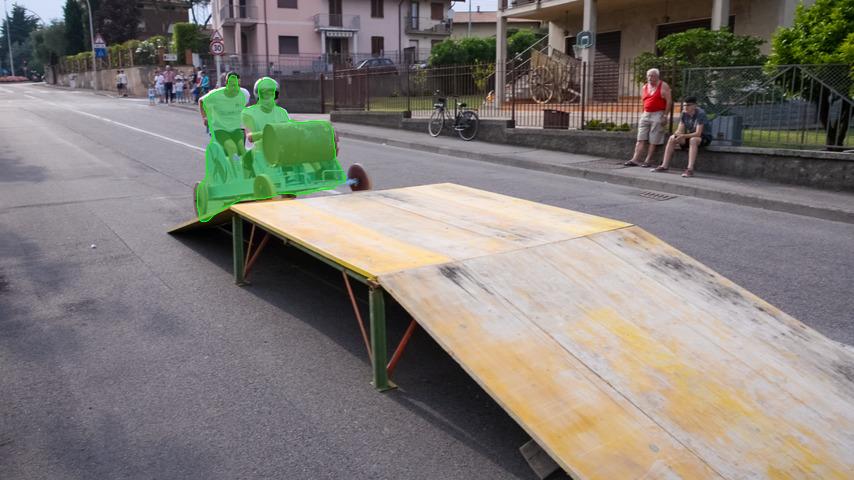}} \hfil  
\subfloat{\includegraphics[width=0.24\linewidth]{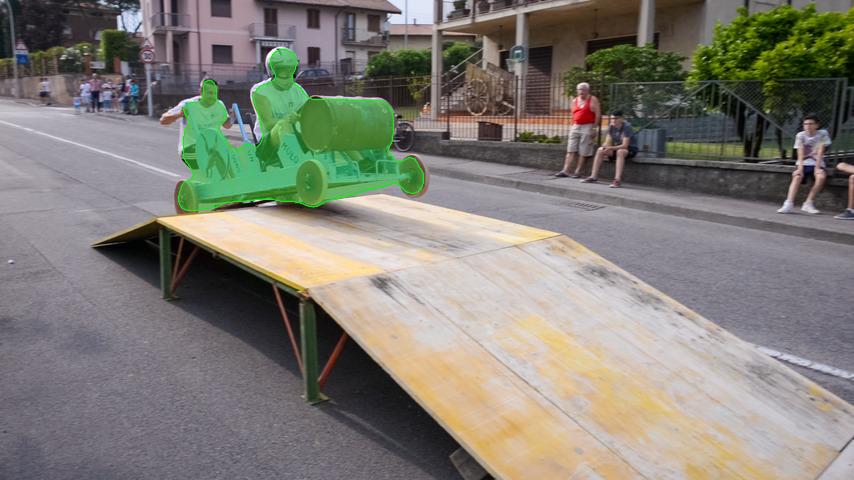}} \hfil  
\subfloat{\includegraphics[width=0.24\linewidth]{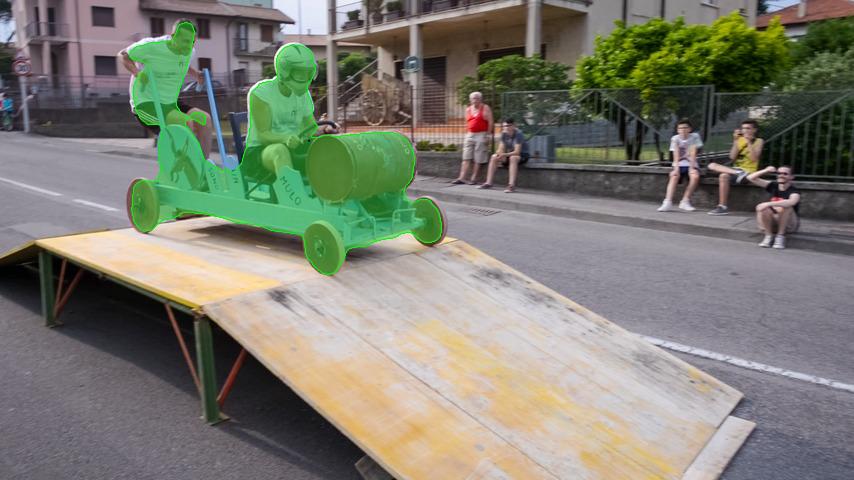}} \hfil \\
\vspace{-0.10in}
\subfloat{\includegraphics[width=0.24\linewidth]{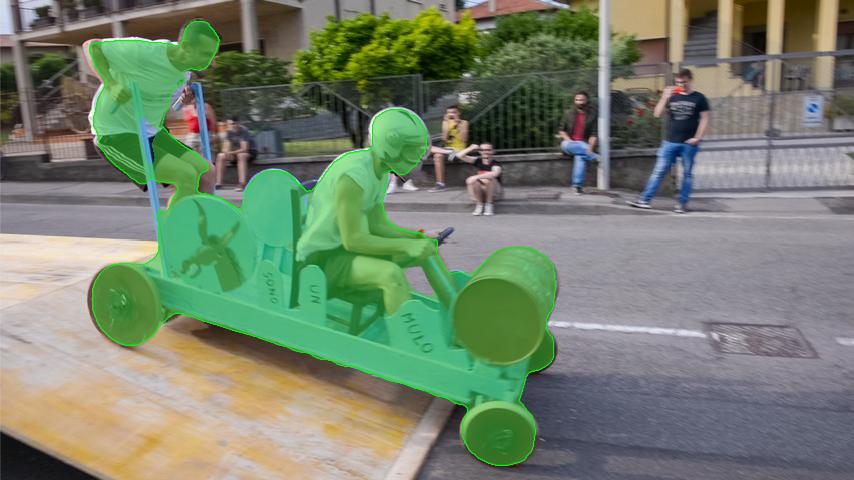}} \hfil 
\subfloat{\includegraphics[width=0.24\linewidth]{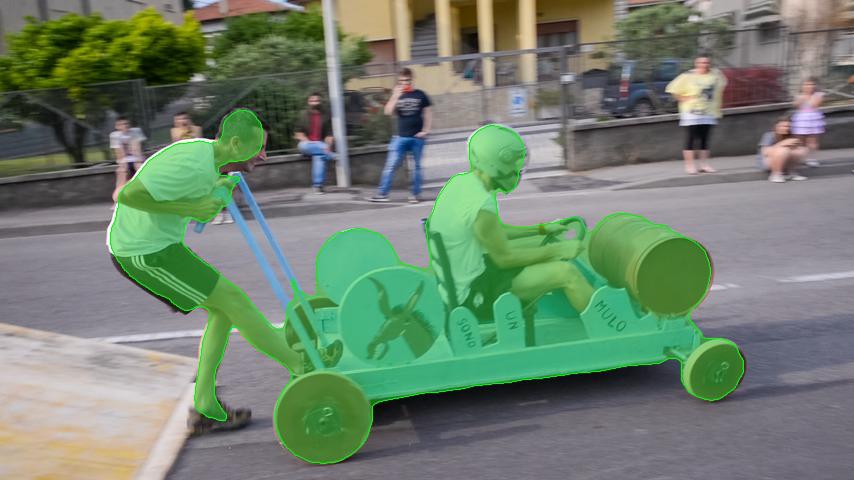}} \hfil 
\subfloat{\includegraphics[width=0.24\linewidth]{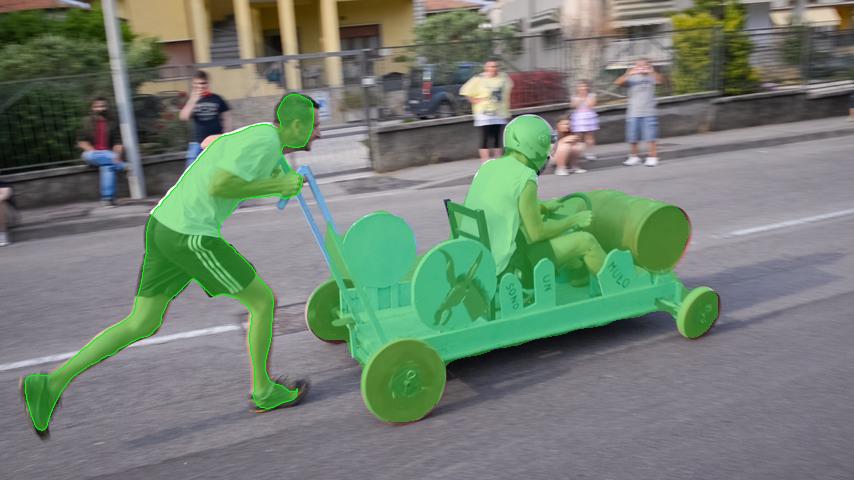}} \hfil 
\subfloat{\includegraphics[width=0.24\linewidth]{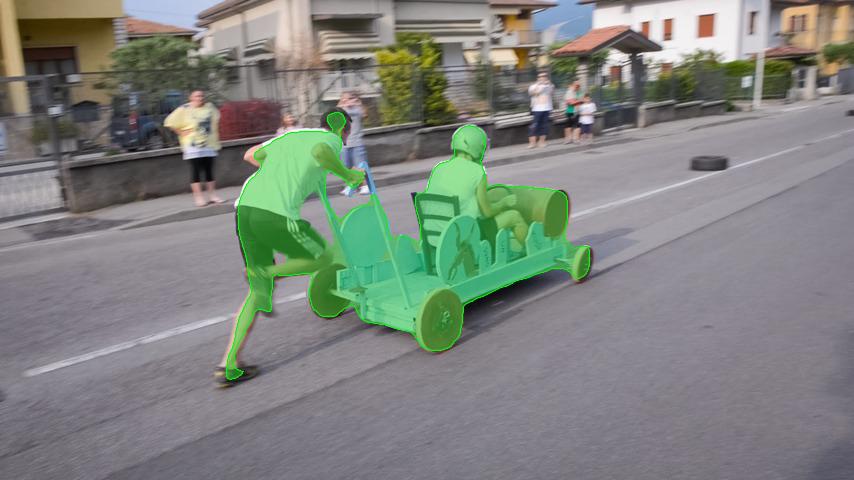}} \hfil 
\caption{Example results of our method, where the pseudo ground truth of
the first frame is in yellow, and the other seven images in green are
sample segmentations of the rest of the video clip. Best viewed in color.}\label{fig:example}
\end{figure}

In this paper, we propose to tag the main object in a video clip by
combining the motion information and the instance segmentation result.
We use optical flow to group segmented pixels to a single object as the
pseudo ground truth and, then, take it as the first frame mask to
perform the OSVOS. The pseudo ground truth is the estimated object mask
for the first frame to replace the true ground truth in the
semi-supervised VOS methods. The main idea is sketched below. We apply a
powerful instance segmentation algorithm, called the Mask R-CNN
\cite{he2017mask}, to the first frame of a video clip as shown in Figure
\ref{fig:tag}, where different objects have different labels. Then, we
extract optical flow from the first two frames and select and group
different instance segmentations to estimate the foreground object.
Next, we finetune a pretrained CNN using the estimated foreground object
from the first frame as the pseudo ground truth and propagate the
foreground/background segmentation to the remaining frames of the video
one frame at a time.  Finally, we achieve state-of-the-art performance
in the benchmark datasets by incorporating online adaptation
\cite{voigtlaender2017online}. Example results are shown in Figure \ref{fig:example}.

\begin{figure*}[t]
\centering
\footnotesize
\includegraphics[width=0.98\linewidth]{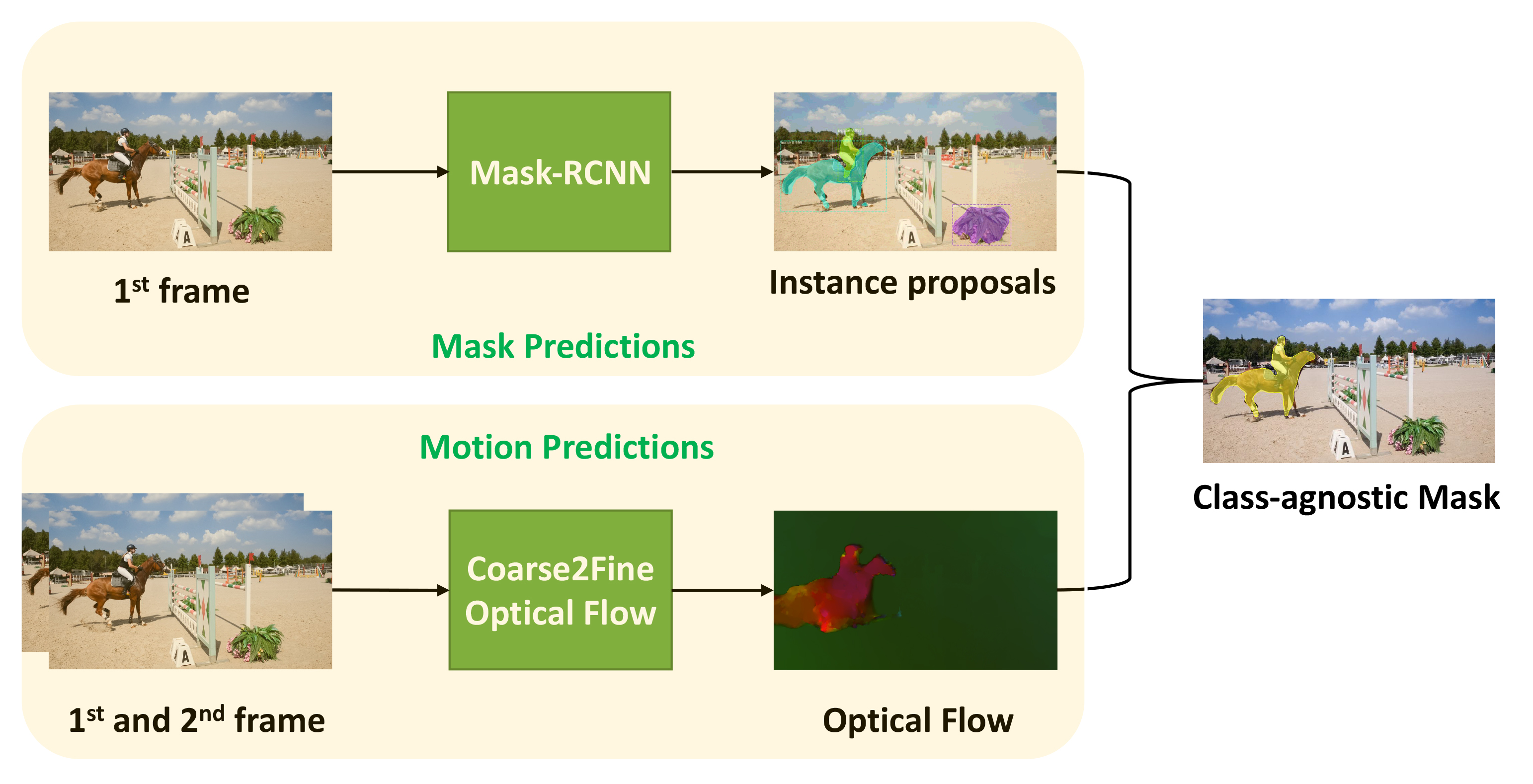}\label{fig:tag}
\caption{Overview of tagging the main object. We use instance
segmentation algorithm to segment objects in the static image. We then
utilize optical flow to select and group the segments to one foreground
object.}\label{fig:tag}
\end{figure*}

Our goal is to segment the primary video object without manual
annotations. The proposed method does not use the temporal information
of the whole video clip at once but one frame at a time. Errors from
each consequent frames do not propagate along time. As a result, the
proposed method has higher tolerance against occlusion and fast motion.
We evaluate the proposed method extensively on the DAVIS dataset \cite{Perazzi2016}, the FBMS dataset \cite{ochs2014segmentation}.  Our method gives state-of-the-art
performance in both datasets with the mean intersection-over-union (IoU)
of 79.3\% on DAVIS, and 77.9\% on FBMS.

Main contributions in this work are summarized below.  First, we
introduce a novel unsupervised video object segmentation method by
combining instance segmentation and motion information.  Second, we
transfer a recent semi-supervised network architecture to the
unsupervised context. Finally, the proposed method outperforms state-of-the-art unsupervised methods on several benchmarks datasets. 

The rest of this paper is organized as follows. Related work is reviewed
in Sec. \ref{sec:rel}.  Our novel unsupervised video object segmentation
method is proposed in Sec. \ref{sec:app}. Experimental results are shown
in Sec. \ref{sec:exp}. Finally, concluding remarks are given in Sec.
\ref{sec:con}. 

\section{Related Work}\label{sec:rel}


\subsubsection{Instance segmentation.} Many video object segmentation
methods \cite{voigtlaender2017online,jain2017fusionseg,perazzi2017learning,caelles2017one} are based on semantic segmentation
networks \cite{wu2016wider} for static images.
State-of-the-art semantic segmentation techniques are dominated by fully
convolutional networks\cite{long2015fully,chen2016deeplab}. Semantic
segmentation segments the same category of objects with one mask while
instance segmentation\cite{he2017mask} provides a segmentation mask
independently for each instance. One key reason that these deep learning
based methods for instance segmentation have developed very rapidly is
that there are large datasets with instance mask annotations such as
COCO \cite{lin2014microsoft}. It is difficult to annotate all categories
of objects and apply a supervised training. It is more difficult to
extend image instance segmentation to video instance segmentation due to
the the lack of large-scale manual labeled instance video object
segmentation datasets. In contrast, we focus on generic object
segmentation in the video and we do not care whether the object category
is in the training dataset or not. We propose a method to transfer the image
instance segmentation to enable finetuning the pretrained fully
convolutional network. 

\subsubsection{Semi-supervised video object segmentation.} Semi-supervised
VOS requires the manual labels for the first frame and then propagate it
to the rest of the video. The annotation provides a good initialization
for the object appearance model and the problem can be considered as a
foreground/background segmentation guided by the first frame annotation.
Deep learning approaches have achieved higher performance
\cite{voigtlaender2017online}, most of the recent work are based on OSVOS
\cite{caelles2017one} and MaskTrack \cite{perazzi2017learning}. OSVOS
creates a new model for each new video initialized with the pretrained
model and finetunes on the first frame without using any temporal
information. OSVOS treats each frame independently while MaskTrack
considers the relationship between consecutive frames when training the
network. Lucid data dreaming \cite{khoreva2017lucid} proposed a data
augmentation technique by cutting-out foreground, in-painting the
background, perturbing both the foreground and background and finally
reconstructed the frames. VOS with re-identification \cite{li2017video}
adds a re-identification step to recover the lost instances in the long
term by feeding the cropped patch contained the object instead of
forwarding the entire image to the network. OnAVOS
\cite{voigtlaender2017online} proposed a online finetuning approach to
segment future frames based on the first frame annotation and the
previous predicted segmented frames. 

\subsubsection{Unsupervised video object segmentation.} Unsupervised VOS
algorithms \cite{jain2017fusionseg,koh2017primary} discover the primary object
segmentation in a video and assume no manual annotations. Some
approaches formulate segmentation as foreground and background labeling
problem, such as Gaussian Mixture Models and Graph Cut
\cite{marki2016bilateral,tsai2016video}. ARP \cite{koh2017primary}
proposed a unsupervised video object segmentation approach by
iteratively refining the augmentation with missing parts or reducing
them by excluding noisy parts. Recently more CNN-based approaches
identify the primary object by using motion boundaries, saliency maps
\cite{tokmakov2017learning,jain2017fusionseg}. LMP
\cite{tokmakov2017learning} trains an encoder-decoder architecture using
ground truth optical flow and motion segmentation and then refines by
the objectness map. Both LVO \cite{tokmakov2017learninglvo} and FSEG
\cite{jain2017fusionseg} have two-stream fully convolutional neural
networks that combine appearance features and motion features, LVO
further improves the segmentation by forwarding the features to
bidirectional convolutional GRU, while FSEG fuses these two models and
put it as an end-to-end training. Unsupervised approach is more desired
since it needs no human interactions and we focus on the unsupervised
VOS in this paper. 

\section{Proposed Method}\label{sec:app}

\begin{figure*}[t]
\centering
\footnotesize
\includegraphics[width=0.98\linewidth]
{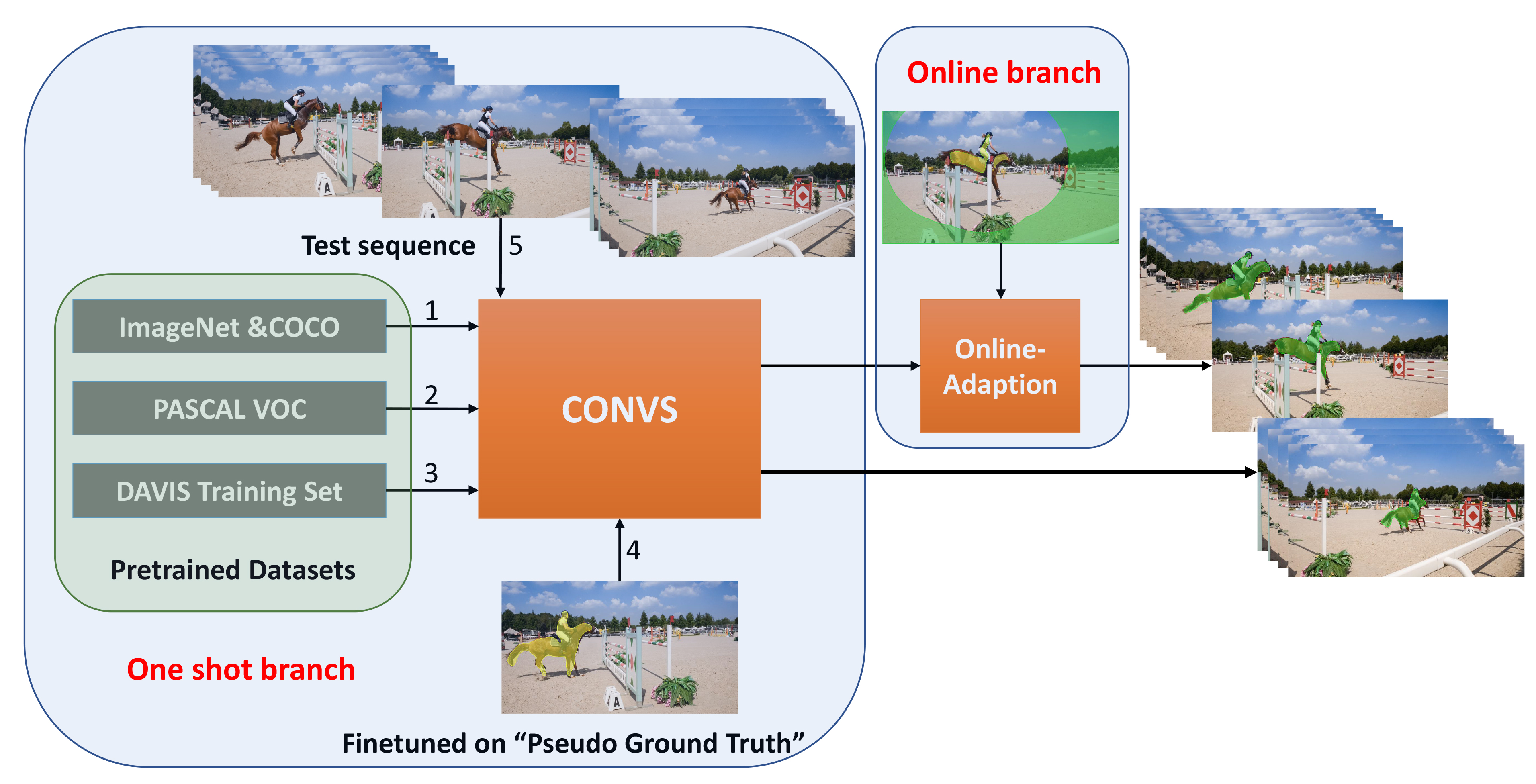}\label{fig:pipeline}
\caption{Overview of the proposed method. We trained the appearance model on the DAVIS training set with a pre-trained wider-ResNet from ImageNet, COCO and PASCAL VOC. We then finetuned the model on the first frame ``pseudo ground truth'', and online adaptation is applied afterwards. The pixels in yellow and green are selected positive and negative examples respectively in the online branch.}\label{fig:pipeline}
\end{figure*}

Our goal is to segment generic object in the video in an unsupervised
approach. In the semi-supervised VOS, the first frame ground truth label
is needed. Inspired by the semi-supervised approach, we propose a method
to tag the ``pseudo ground truth'' and then take it as input for the
pretrained network, and then output the segmentation masks for the rest
of the video. To our best knowledge, this is the first attempt to transfer semi-supervised VOS approach to unsupervised VOS approach by utilizing ``pseudo ground truth''. Figure \ref{fig:pipeline} shows the overview of the proposed method, which includes 
three key components, the criterion to tag the primary object, appearance model and online adaptation.

\subsection{Learning to tag the foreground object}\label{sec:lea}

\subsubsection{Image instance segmentation.} We apply an image-based instance
segmentation algorithm to the first frame of the given video.
Specifically, we choose Mask R-CNN \cite{he2017mask} as our instance
segmentation framework and generate instance masks. We further exploit
the error analysis to demonstrate that better initial instance
segmentations improve the performance in a large margin which suggests
that our proposed method has the potential to improve further with more
advanced instance segmentation methods. 

Mask R-CNN is a simple yet high performance instance segmentation model.
Specifically, Mask R-CNN adds an additional FCN mask branch to the
original Faster R-CNN \cite{ren2015faster} model. The mask branch
and the bounding box branch are trained simultaneously in the training, while
the instance masks are generated from the detection results
at inference time. The box prediction branch generates bounding boxes based on the  proposals followed by non-maximum suppression. The mask branch is then applied to predict segmentation masks from the 100 detection boxes with the highest scores. This step speeds up the inference time and improves accuracy, which is different from the training step with parallel computation. For each region of interests (ROIs), the mask can predict \emph{n} times where \emph{n} is the class number in the training set, and the only used \emph{k}-th mask is from the predicted class by the classification branch. 

We note that the mask branch generates class-specific
instance segmentation masks for the given image whereas VOS focuses on
class-agnostic object segmentation. Our experiments show that even
though Mask R-CNN can only generate limited-class labels due to the
labels of COCO \cite{lin2014microsoft} and PASCAL
\cite{everingham2010pascal}, we can still output instance segmentation
masks with closest class label. Our algorithm needs to further force all
the classes to one foreground class, and thus the mis-classification has
little influence to the performance of VOS. 

\noindent
\subsubsection{Optical flow thresholding.} There are two important cues in video object segmentations: appearance and motion. To use the information from both spatial and temporal domain, we incorporate optical flow with instance segmentation to learn to segment the primary object. Instance segmentation can generate precise
class-specific segmentation masks, however, the algorithm cannot
determine the primary object in the video. While optical flow can separate moving objects from the background, however the optical flow esimation is still far from perfect. Motivated by the moving objects attract
people's attention \cite{pathak2017learning}, so we use motion
information to select and group the static image instance segmentation
proposals, which takes advantange of the merits of optical flow and instance segmentation. We apply optical flow algorithm Coarse2Fine \cite{liu2009beyond} to
extract the optical flow between the first frame and the second frame of
a given video clip. To combine with the instance segmentation proposals, we
normalize the flow magnitude and then threshold and select the optical
flow motivated by the faster motions are more likely to attract
attentions. 

We select instance segmentation proposals which have more than 80\%
overlap with optical flow mask. We further group the selected proposal
masks with different class labels to one foreground class without any
class labels. In image-based instance segmentation, the same object may
be separated into different parts due to the differences of colors,
textures and the influence of occlusions. We can efficiently group the
different parts to one primary object without knowing the categories of
the objects. We named this foreground object as ``pseudo ground truth''
and forward it to the pretrained appearance model. Sample
``pseudo ground truths'' are shown in Figure \ref{fig:tag}. 

\subsection{Unsupervised video object segmentation}\label{sec:uns}

Our proposed method is built on one-shot video object segmentation
(OSVOS) \cite{caelles2017one} which finetunes the pretrained appearance
model on the first annotated frame. We replace the first annotated
frame with our estimated ``pseudo ground truth'' so that semi-supervised
network architecture can be used in our proposed approach. Our goal is
to train a ConvNet to segment a generic object in a video. 

\noindent
\subsubsection{Network overview.} We adopt a more recent ResNet \cite{wu2016wider}
architecture pretrained on ImageNet \cite{deng2009imagenet} and MS-COCO
\cite{lin2014microsoft} to learn powerful features. In more detail, the network uses the model A of the wider ResNet with 38 hidden layers as the backbone. The data in DAVIS
training datasets is very scarce and we further pretrain the network
using PASCAL \cite{everingham2010pascal} by mapping all the 20-class
labels to one foreground label and keep background unchanged. As
demonstrated in OnAVOS \cite{voigtlaender2017online}, the two steps of
finetuning on DAVIS and PASCAL are complementary. Hence, we finetune the
network using DAVIS training datasets and obtain the final pretrained
network. The above training steps are all offline training to construct
a model to identify foreground object. At inference time, we finetune
the network on the ``pseudo ground truth'' of the first frame to
tell the network which object to be segmented. However, the first frame does
not provide all the information through the whole video, and thus online
adaptation is needed during the test time. 

\noindent
\subsubsection{Online adaptation.} The major difficulty for video object segmentation is the appearance may change dramatically throughout the video. A model learned only from the first frame cannot address the severe appearance changes. Therefore online adaptation for the model is needed to exploit the information from the rest frames during inference time. 

We adopt test data augmentation method from
Lucid Data Dreaming augmentation \cite{khoreva2017lucid} and online
adaptation approach from OnAVOS \cite{voigtlaender2017online} to perform
our online finetuning. We generate augmentation of the first frame using
Lucid Data Dreaming approach. As each frame is segmented, foreground
pixels with high confidence predictions are taken as further positive
training examples, while pixels far away from the last assumed object
position are taken as negative examples. Then an additional round of
fine-tuning is performed on the newly acquired data. 

\section{Experiments}\label{sec:exp}

To evaluate the effectiveness of our proposed method, we conduct
experiments on three challenging video object segmentation datasets:
DAVIS \cite{Perazzi2016}, Freiburg-Berkeley Motion Segmentation (FBMS)
dataset \cite{ochs2014segmentation}, SegTrack-v2 \cite{li2013video}. We
use region similarity, which is defined as the intersection-over-union
(IoU) between the estimated segmentation and the ground truth mask, and F-score evaluation protocol proposed in \cite{ochs2014segmentation} to estimate the accuracy. 

\subsection{Datasets}

We provide a detailed introduction to evaluation benchmarks below. 

\noindent
\subsubsection{DAVIS.} The DAVIS dataset is composed of 50 high-definition video
sequences, 30 in the training set and the remaining 20 in the validation set. There are totally 3, 455 densely annotated, pixel-accurate frames.  The videos contain challenges such as occlusions, motion blur, and appearance
changes. Only the primary moving objects are annotated in the
ground truth. 

\noindent
\subsubsection{FBMS.} The Freiburg-Berkeley motion segmentation dataset is
composed of 59 video sequences with 720 frames annotated. In contrast to
DAVIS, it has multiple moving objects in several videos with instance-level annotations. We do not train on any of these sequences and evaluate using mIoU and F-score respectively. We also
convert the instance-level annotations to binary ones by merging all
foreground annotations, as in \cite{tokmakov2017learning}. 

\noindent
\subsubsection{SegTrack-v2.} The SegTrack-v2 dataset contains 14 videos with a
total of 1, 066 frames with pixel-level annotations. For videos with
multiple objects with individual ground-truth segmentations, each object can be segmented in turn, treating each as a problem of segmenting that object from the background.

\begin{figure}[t]
\captionsetup[subfigure]{}
\centering 
\subfloat{\includegraphics[width=0.187\linewidth]{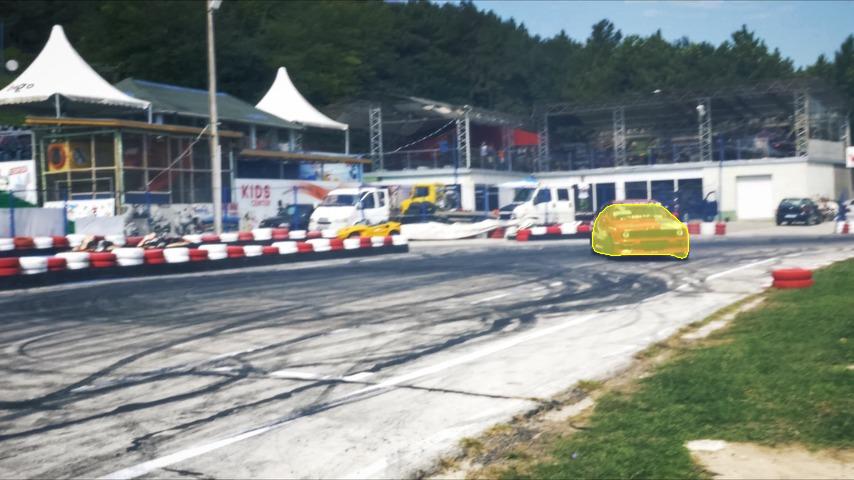}} \hfil
\subfloat{\includegraphics[width=0.187\linewidth]{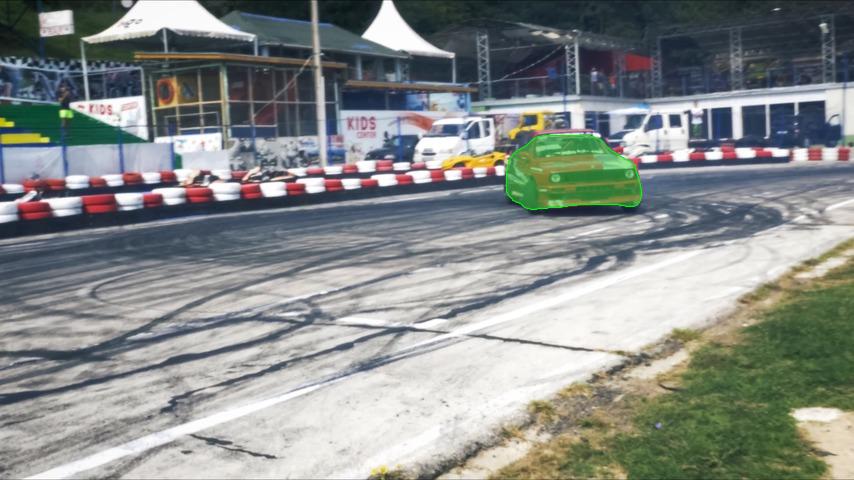}} \hfil
\subfloat{\includegraphics[width=0.187\linewidth]{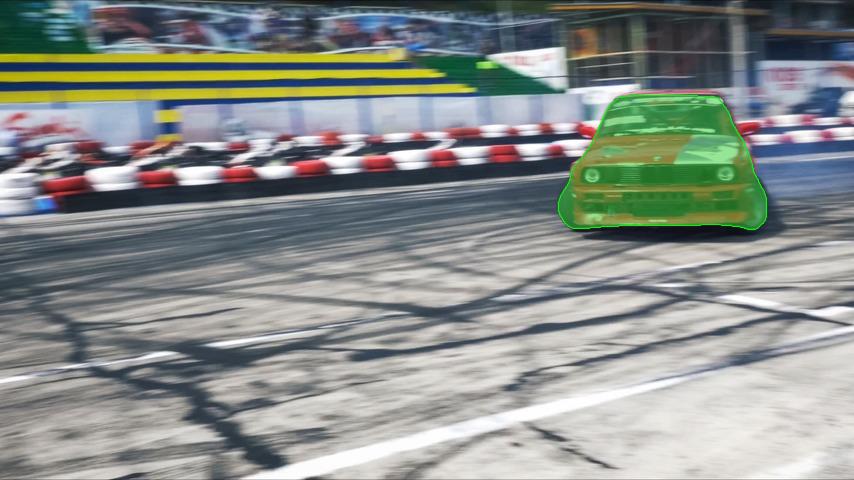}} \hfil
\subfloat{\includegraphics[width=0.187\linewidth]{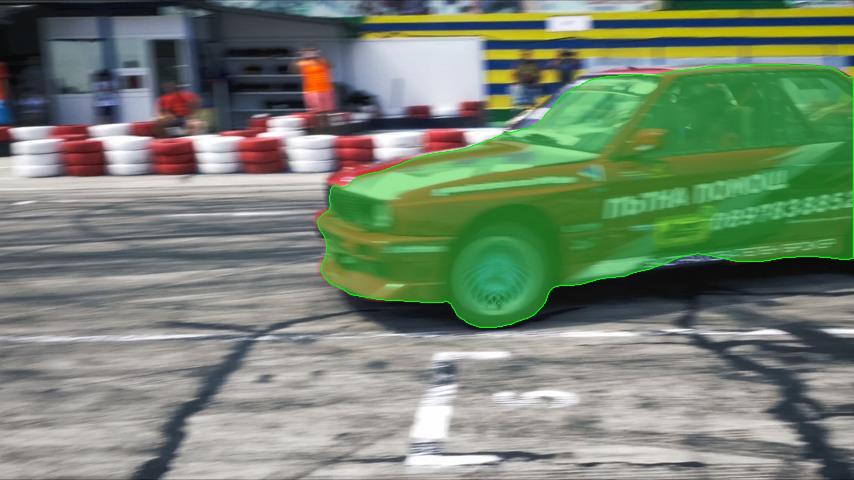}} \hfil
\subfloat{\includegraphics[width=0.187\linewidth]{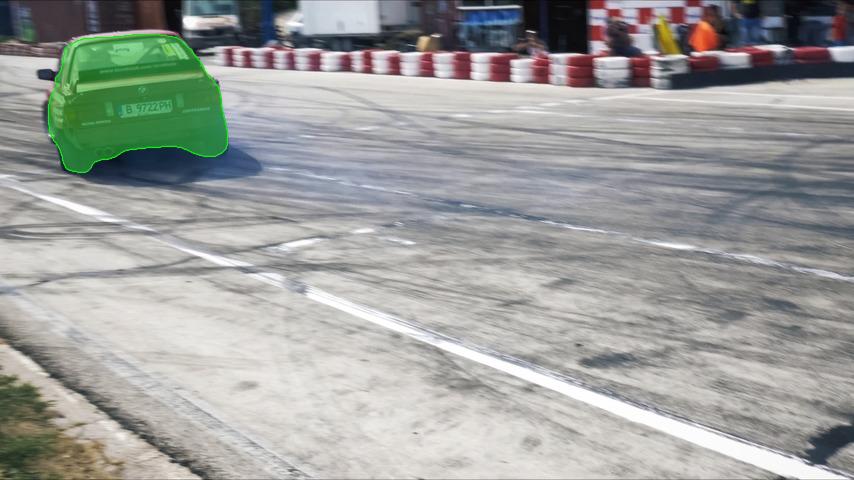}} \hfil \\
\vspace{-0.1in}
\subfloat{\includegraphics[width=0.187\linewidth]{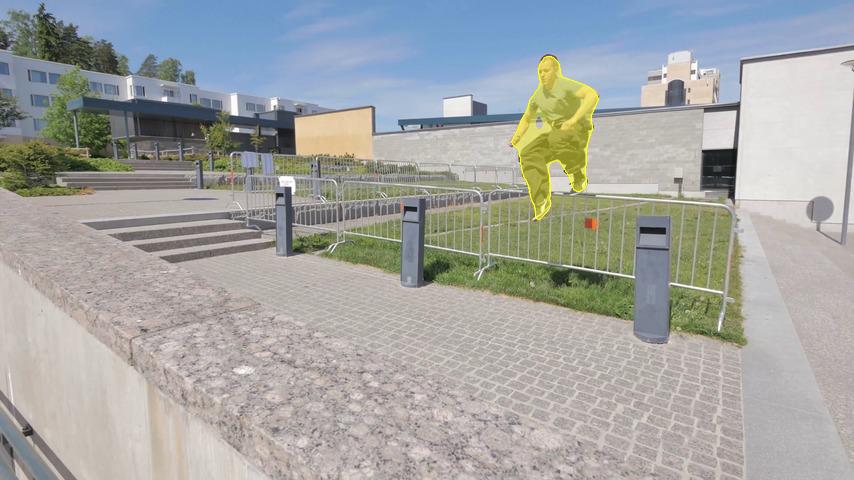}} \hfil
\subfloat{\includegraphics[width=0.187\linewidth]{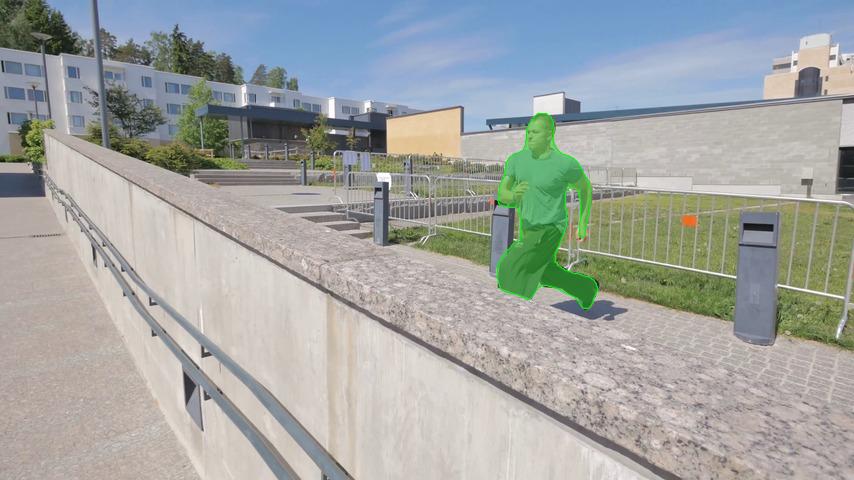}} \hfil
\subfloat{\includegraphics[width=0.187\linewidth]{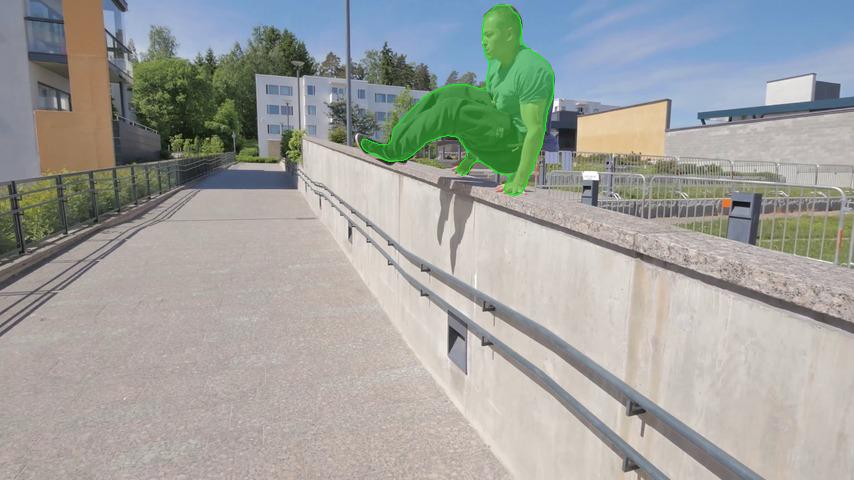}} \hfil
\subfloat{\includegraphics[width=0.187\linewidth]{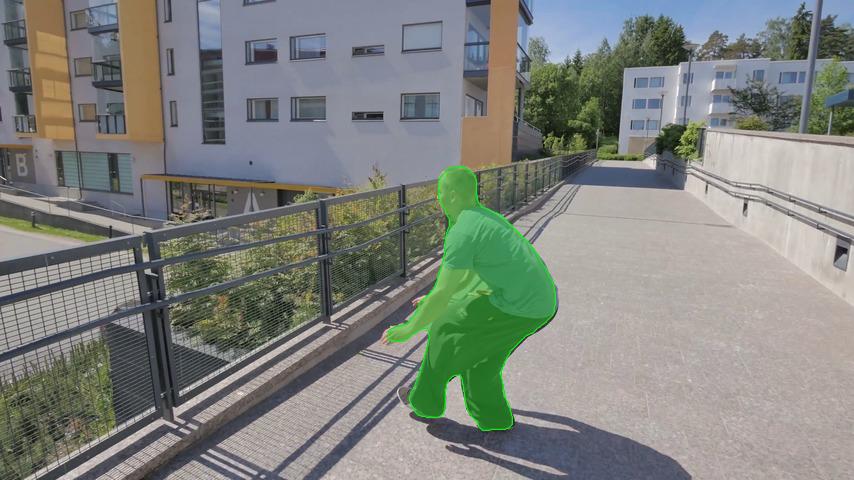}} \hfil
\subfloat{\includegraphics[width=0.187\linewidth]{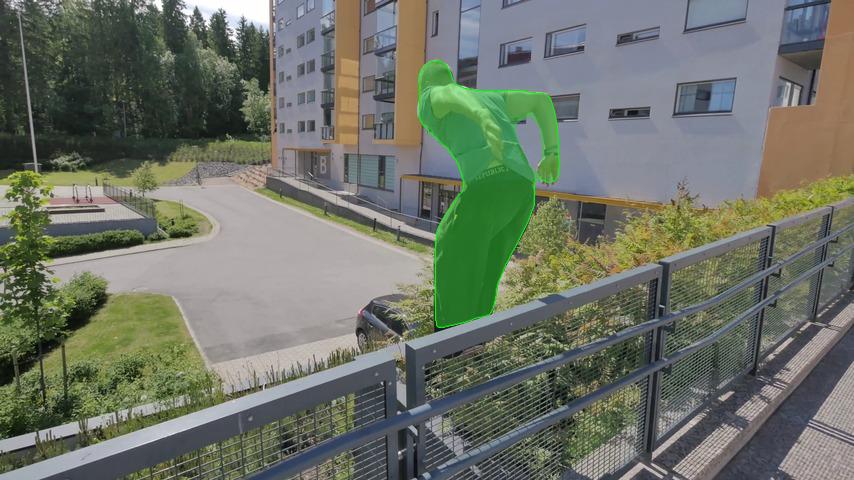}} \hfil \\
\vspace{-0.1in}
\subfloat{\includegraphics[width=0.187\linewidth]{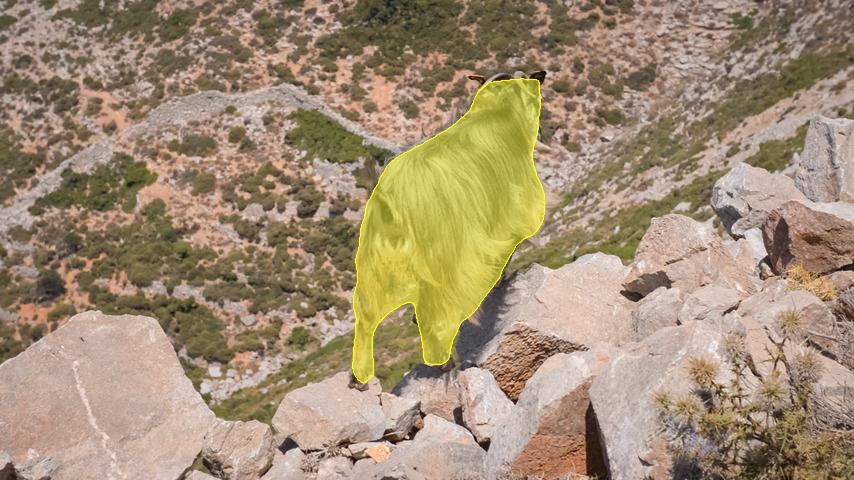}} \hfil
\subfloat{\includegraphics[width=0.187\linewidth]{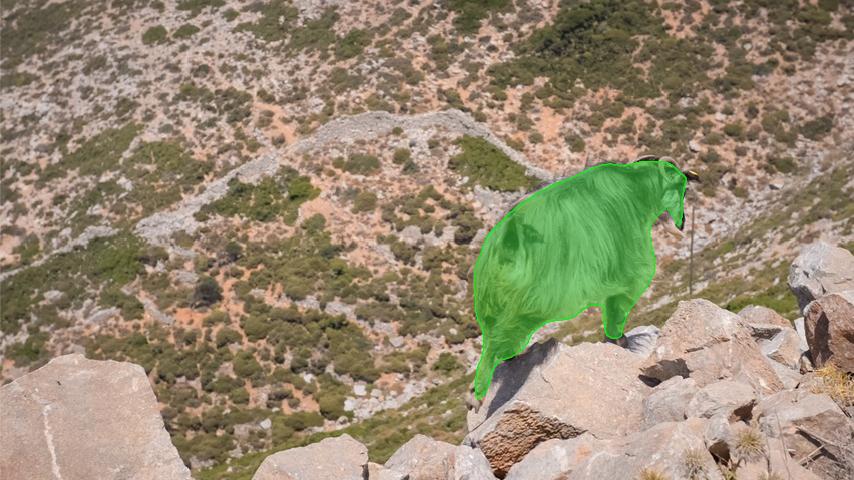}} \hfil
\subfloat{\includegraphics[width=0.187\linewidth]{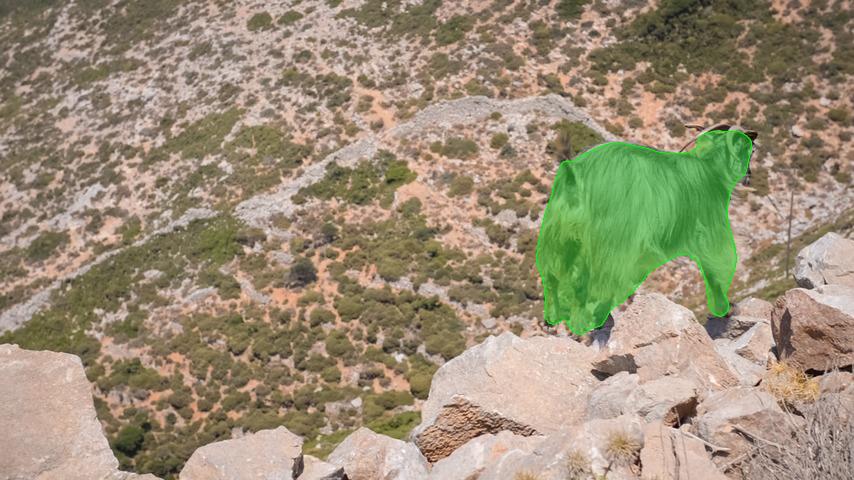}} \hfil
\subfloat{\includegraphics[width=0.187\linewidth]{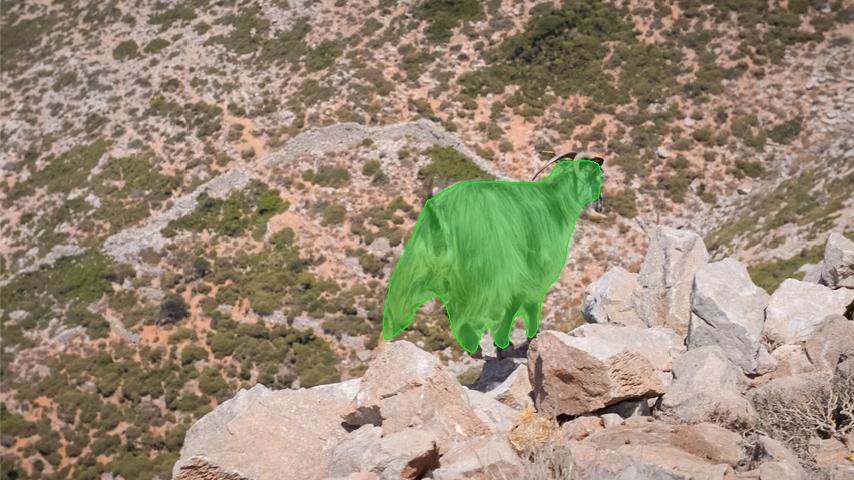}} \hfil
\subfloat{\includegraphics[width=0.187\linewidth]{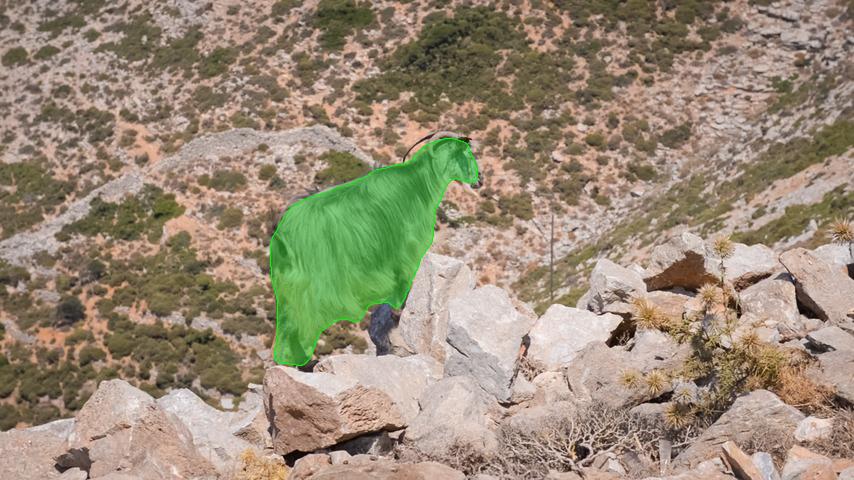}} \hfil\\
\vspace{-0.1in}
\subfloat{\includegraphics[width=0.187\linewidth]{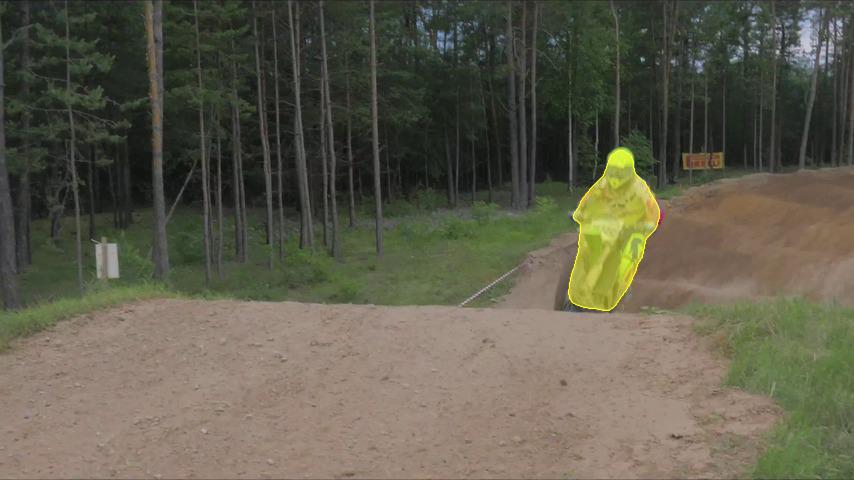}} \hfil
\subfloat{\includegraphics[width=0.187\linewidth]{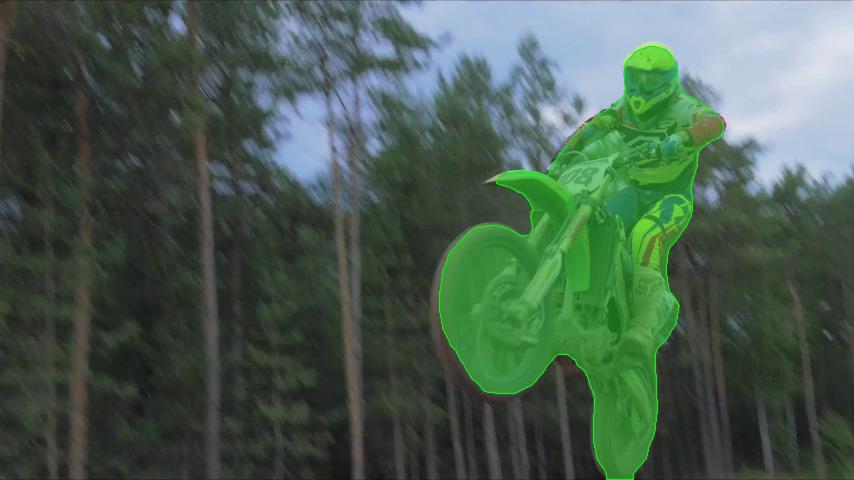}} \hfil
\subfloat{\includegraphics[width=0.187\linewidth]{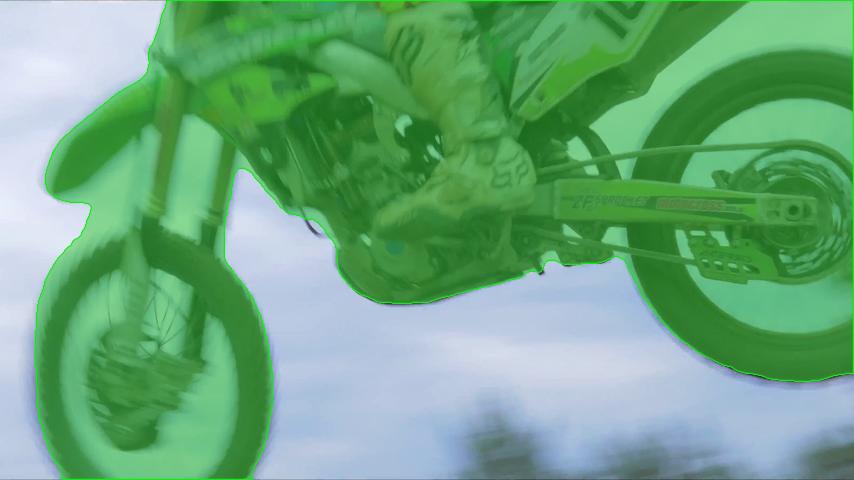}} \hfil
\subfloat{\includegraphics[width=0.187\linewidth]{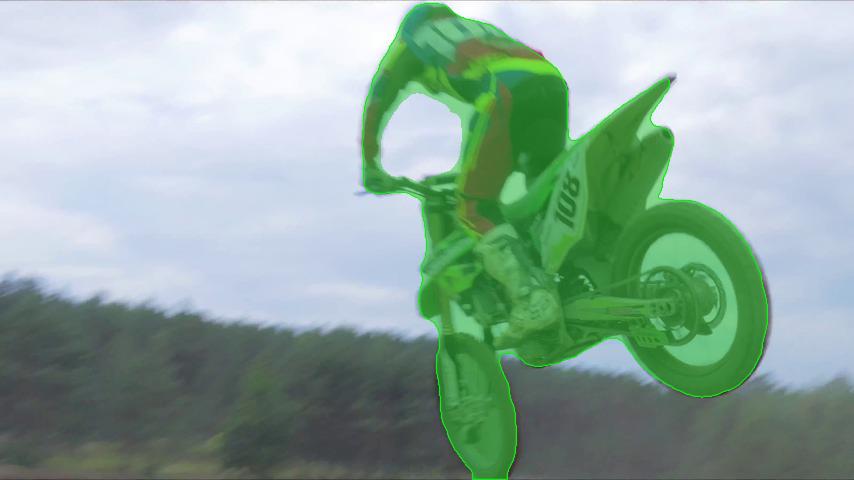}} \hfil
\subfloat{\includegraphics[width=0.187\linewidth]{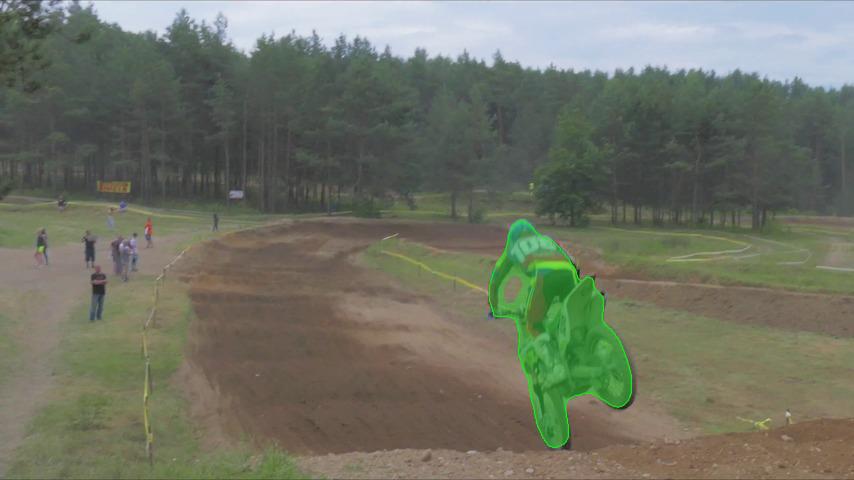}} \hfil\\
\vspace{-0.1in}
\subfloat{\includegraphics[width=0.187\linewidth]{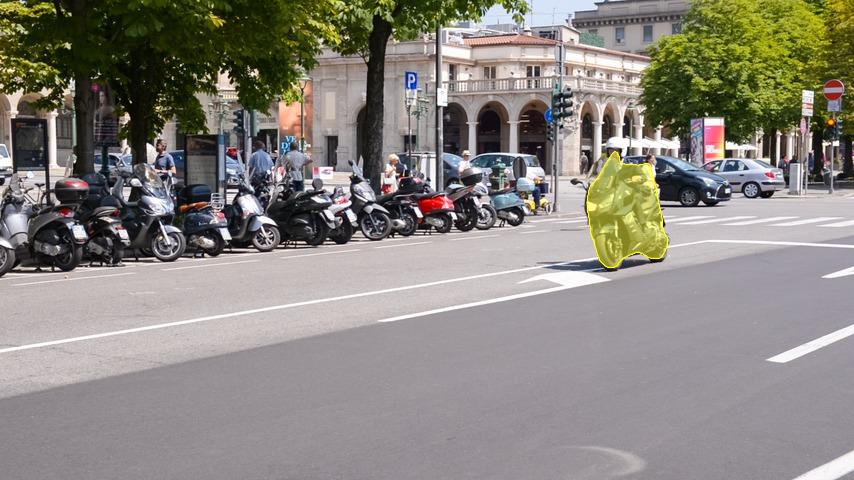}} \hfil
\subfloat{\includegraphics[width=0.187\linewidth]{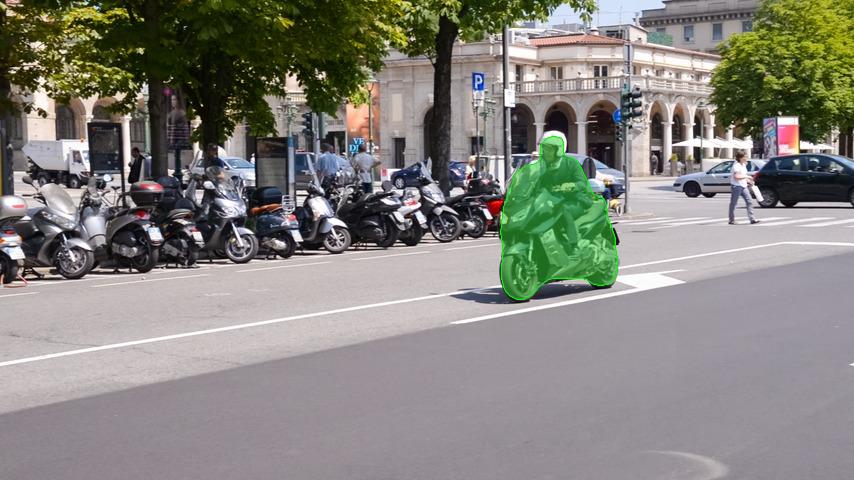}} \hfil
\subfloat{\includegraphics[width=0.187\linewidth]{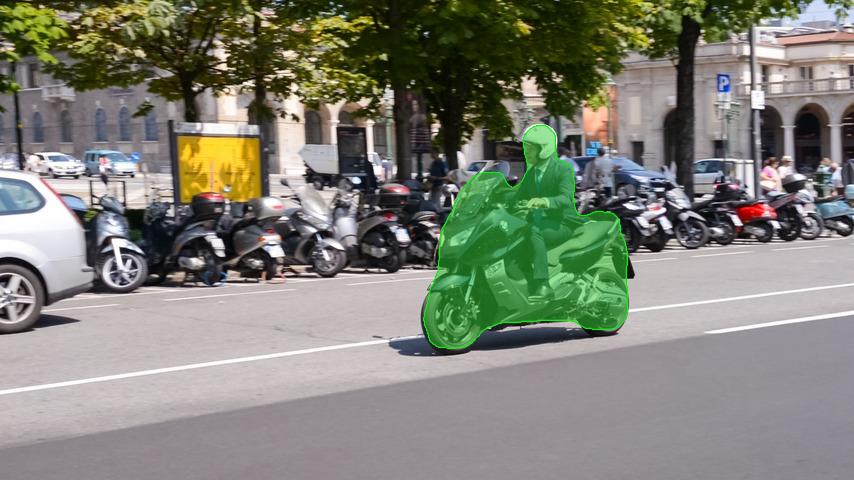}} \hfil
\subfloat{\includegraphics[width=0.187\linewidth]{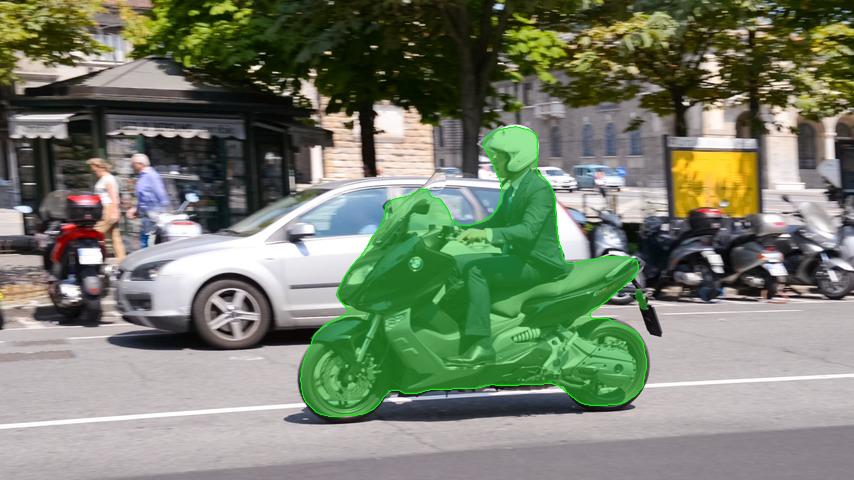}} \hfil
\subfloat{\includegraphics[width=0.187\linewidth]{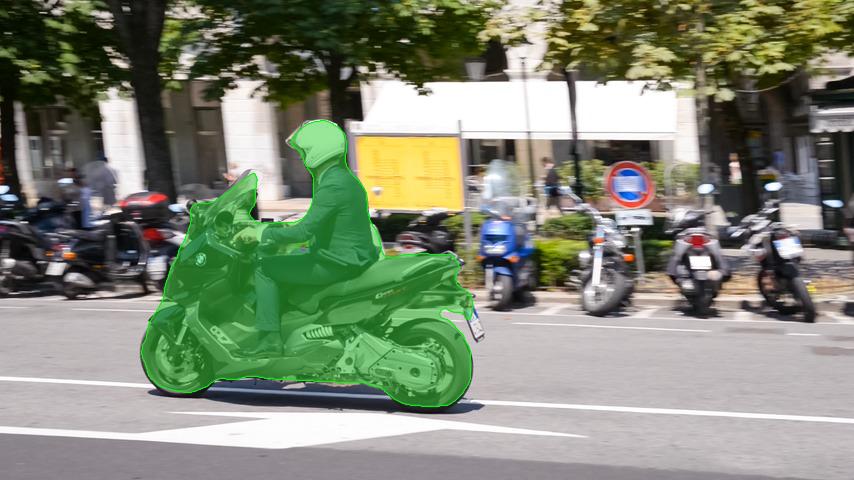}} \hfil\\
\caption{Qualitative results on DAVIS validation set: The first column
in yellow is the ``pseudo ground truth'' of the first frame of each
video. The other four columns are the output segmentation masks of our
proposed approach. Our algorithm performs well on videos with fast motion (first and forth row), gesture changes (second row), unseen category (third row) and complex background (fifth row). Best viewed in color.}\label{fig:exp}
\end{figure}

\begin{table}[t]
\centering
\caption{Comparison of the mIoU scores ($\%$) of different unsupervised
VOS approaches in the DAVIS 2016 validation dataset. Our method achieves
the highest mIoU compared with state-of-the-art methods} 
\label{tab:davisfull}
\begin{tabular}{|c|c|c|c|c|c|c||c|}
\hline
 & NLC \cite{faktor2014video} & FST \cite{papazoglou2013fast} & LMP \cite{tokmakov2017learning} & FSEG \cite{jain2017fusionseg} & LVO \cite{tokmakov2017learninglvo} & ARP \cite{koh2017primary} & Ours \\  \hline \hline
mIoU & 55.1 & 55.8 & 70.0 & 70.7 & 75.9 & 76.2 & \bf{79.3} \\ \hline
\end{tabular}
\end{table}

\subsection{Implementation details}

We jointly use optical flow and semantic instance segmentation to group
foreground objects that move together into a single object. We use the
optical flow from a re-implementation of Coarse2Fine optical flow
\cite{liu2009beyond}. We implemented the objectness network using
Tensorflow \cite{abadi2016tensorflow} library and set wider ResNet \cite{wu2016wider} with 38 hidden layers as the backbone. The segmentation network is simple without using upsampling, skip connections or multi-scale structures. In some convolution layers, increasing the dilation rates and removing the down-sampling operations accordingly are applied to generate score maps at 1/8 resolution. Large field-of-view setting in Deeplabv2 \cite{chen2016deeplab} is used to replace the top linear classifier and global pooling layer which exist in the classification network. Besides, the batch normalization layers are freezed during finetuning.

We adopted the initial network weights provided by the
repository which were pre-trained on the ImageNet and COCO dataset. We
further finetune the objectness network based on the augmented PASCAL
VOC ground truth from \cite{hariharan2014simultaneous} with a total of
12, 051 training images. Note that we force all the foreground objects
in a certain image to one single foreground object and keep background the same.

For the DAVIS dataset evaluation, we further train the network on DAVIS
training set and then apply a one-shot finetuning on the first frame
with ``pseudo ground truth''. The segmentation network is trained on the first frame image/``pseudo ground truth'' pair, by Adam with learning rate $3 \times 10^{-6}$. We set the number of finetuning n$_{f}$ on 
the first frame as 100, we found that a relative small n$_{f}$ can
improve the accuracy which is opposite with semi-supervised VOS. For the
online part, we used the default parameters in OnAVOS
\cite{voigtlaender2017online} by setting the number of finetuning as 15, finetuning interval as 5 frames, and learning rate as $1 \times 10^{-5}$ and adopted the CRF parameters from
DeepLab \cite{chen2016deeplab}. For completeness, we also conduct
experiments on FBMS and SegTrack-v2 datasets, we conduct the same
procedures for FBMS as DAVIS. To check the effectiveness of the ``pseudo
ground truth'' we only perform one-shot branch for SegTrack-v2 without online adaption.

\subsection{Comparison with state-of-the-art methods}

\subsubsection{DAVIS.} We compare our proposed approach with state-of-the-art
unsupervised techniques, NLC \cite{faktor2014video}, LMP
\cite{tokmakov2017learning}, FSEG \cite{jain2017fusionseg}, LVO
\cite{tokmakov2017learninglvo}, and ARP \cite{koh2017primary} in Table \ref{tab:davisfull}. We achieve the best performance for
unsupervised video object segmentation: 3.1\% higher than the second
best ARP. Besides, we achieve mIoU of 71.2\% on the DAVIS validation set by extracting the pseudo ground-truth on each frame of a given video. When we break down the performance on each DAVIS sequence, we
outperform the majority of the videos shown in Table \ref{tab:davis},
and especially for drift-straight, libby and scooter-black, our results
are more than 10\% higher than the second best results. Our approach
could segment unknown object classes which do not need to be in the
PASCAL/COCO vocabulary. The goat in the third row is an unseen category
in the training data, the closest semantic category horse is matched
instead. Note that our algorithm only needs the foreground mask without
knowing the specific category, and performs better than
state-of-the-art methods. Our method performs even better when the
object classes are in the MS COCO, the top two rows show a single
instance segmentation with large appearance changes (first row) and
viewing angle and gesture changes (second row). The bottom two rows show
that our algorithm works well when merging multiple object masks to one
single mask with viewing angle changes (forth row) and messy background
(fifth row). 

To verify where the improvements come from, we utilize similar backbone with previous method. We test OSVOS \cite{caelles2017one} by replacing the first frame annotations with pseudo
ground truths. OSVOS uses the VGG architecture, and we set the number of
first-frame fine-tuning to 500 without applying boundary snapping. The
mIoUs of our approach and the original OSVOS are 72.3\% and 75.7\%,
respectively. Our approach in the VGG architecture still outperforms
FSEG (70.7\%) without online adaptation, CRF, test time data
augmentation.

\begin{table}[t]
\centering
\caption{Comparison of the mIoU scores ($\%$) per video of several
methods for the DAVIS 2016 validation dataset. (1) blackswan, (2)
bmx-trees, (3) breakdance, (4) camel, (5) car-roundabout, (6)
car-shadow, (7) cows, (8) dance-twirl, (9) dog, (10) drift-chicane, (11)
drift-straight, (12) goat, (13) horsejump-high, (14) kite-surf, (15)
libby, (16) motocross-jump, (17) paragliding-launch, (18) parkour, (19)
scooter-black, (20) soapbox}\label{tab:davis}
\resizebox{\textwidth}{!}{%
\begin{tabular}{|c|cccccccccccccccccccc||c|}
\hline
Method & No.1 & No.2 & No.3 & No.4 & No.5 & No.6 & No.7 & No.8 & No.9 & No.10 & No.11 & No.12 & No.13 & No.14 & No.15 & No.16 & No.17 & No.18 & No.19 & No.20 & meanIoU \\  \hline \hline
FSEG \cite{jain2017fusionseg} & 81.2 & 43.3 & 51.2 & 83.6 & 90.7 & 89.6 & 86.9 & 70.4 & 88.9 & 59.6 & 81.1 & 83.0 & 65.2 & 39.2 & 58.4 & 77.5 & 57.1 & 76.0 & 68.8 & 62.4 & 70.7 \\ \hline
ARP \cite{koh2017primary} & \bf{88.1} & \bf{49.9} & \bf{76.2} & \bf{90.3} & 81.6 & 73.6 & \bf{90.8} & \bf{79.8} & 71.8 & 79.7 & 71.5 & 77.6 & \bf{83.8} & 59.1 & 65.4 & 82.3 & \bf{60.1} & 82.8 & 74.6 & 84.6 & 76.2 \\ \hline
Ours-oneshot & 83.3 & 39.8 & 50.3 & 76.1 & 82.9 & 91.9 & 87.5 & 77.3 & \bf{90.1} & \bf{86.1} & 85.4 & 85.1 & 74.1 & 60.1 & 75.5 & 75.5 & 57.3 & \bf{89.9} & 72.7 & 74.9 & 75.8 \\ \hline
Ours-online & 82.0 & 46.0 & 60.7 & 75.5 & \bf{93.0} & \bf{94.6} & 87.6 & 78.9 & 89.3 & 82.9 & \bf{91.7} & \bf{85.7} & 76.8 & \bf{60.3} & \bf{76.0} & \bf{84.7} & 56.9 & \bf{89.9} & \bf{88.1} & \bf{85.1} & \bf{79.3} \\ \hline
\end{tabular}}
\end{table}

We further analyze the finetuning times on the first frames for both semi-supervised and unsupervised approaches in Table \ref{tab:finetune}. In the table, the second column shows that the performance improves with the increasing finetuning times for semi-supervised approach in terms of mIoU, which indicates more finetuning times with image/ground truth pairs can predict better results. The right two columns show the different relationships between the performance in mIoU and finetuning times on the first frames for unsupervised approach. They both achieve the highest performance by setting the number of finetuning as 100, which indicates the model learns better with an appropriate number of finetuning since the pseudo ground truth is not as accurate as the ground truth.

\begin{table}[t]
\centering
\caption{Comparison of mIoU scores ($\%$) of different finetuning times on the first frames of the DAVIS validation set} 
\label{tab:finetune}
\begin{tabular}{|c|c|c|c|}
\hline
 Finetuning times& Semi-supervised oneshot & Unsupervised oneshot & Unsupervised online \\  \hline \hline
50 & 80.4 & 73.1 & 77.6 \\ \hline
100 & 80.7 & {\bf75.8} & {\bf79.3}\\ \hline
500 & 81.4 & 74.4 & 77.7 \\ \hline
2000 & {\bf82.1} & 74.8 & 77.9\\ \hline

\end{tabular}
\end{table}

\noindent
\subsubsection{FBMS.} We evaluate the proposed approach on the test set, with 30
sequences in total. The results are shown in Table \ref{tab:FBMS}. Our method is outperformed in both evaluation metrics, with an F-score of 85.1\% which is 7.3\% higher than the second best method LVO \cite{tokmakov2017learninglvo},
and the mIoU of 77.9\% which is 18.1\% better than ARP
\cite{koh2017primary}, which performs the second best on DAVIS. Figure
\ref{fig:exp} shows qualitative results of our method, our algorithm
performs well for most of the sequences. The last row shows the failure case for rabbits04 since there are severe occlusions in this video and the rabbit is also an unseen category in the MS COCO. To recover a better prediciton mask, further motion information should be used to address this problem.

\begin{figure}[t]
\captionsetup[subfigure]{}
\centering 
\subfloat{\includegraphics[width=0.187\linewidth]{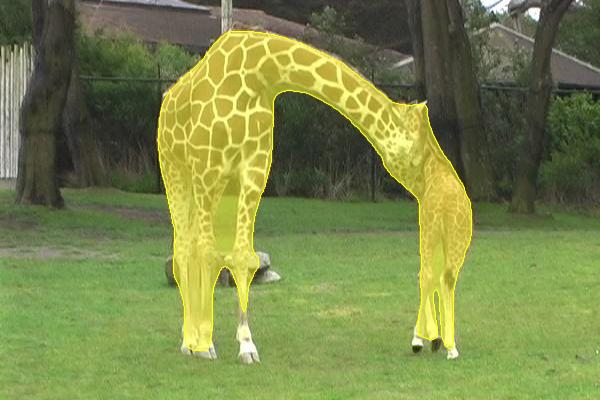}} \hfil
\subfloat{\includegraphics[width=0.187\linewidth]{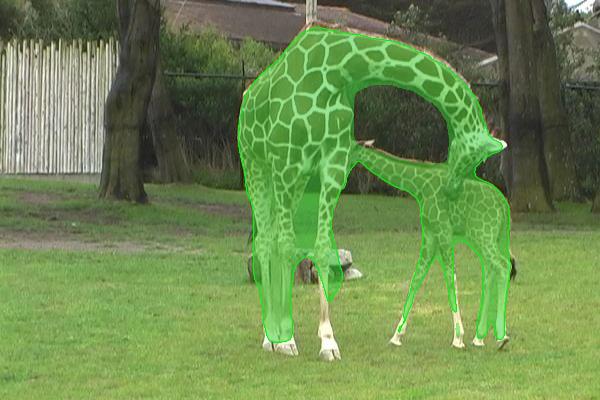}} \hfil
\subfloat{\includegraphics[width=0.187\linewidth]{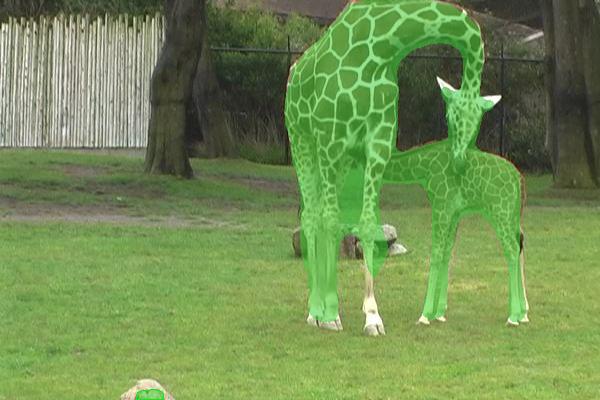}} \hfil
\subfloat{\includegraphics[width=0.187\linewidth]{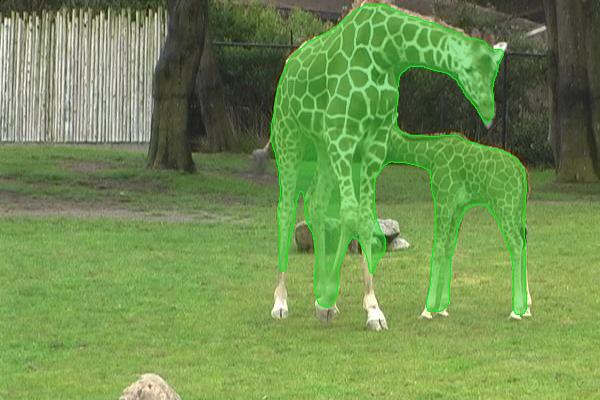}} \hfil
\subfloat{\includegraphics[width=0.187\linewidth]{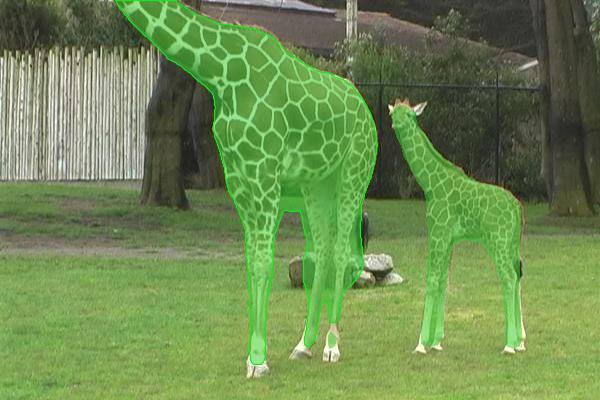}} \hfil \\
\vspace{-0.1in}
\subfloat{\includegraphics[width=0.187\linewidth]{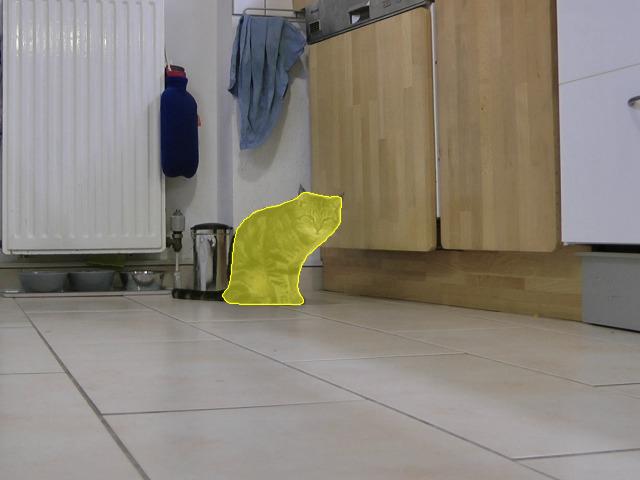}} \hfil
\subfloat{\includegraphics[width=0.187\linewidth]{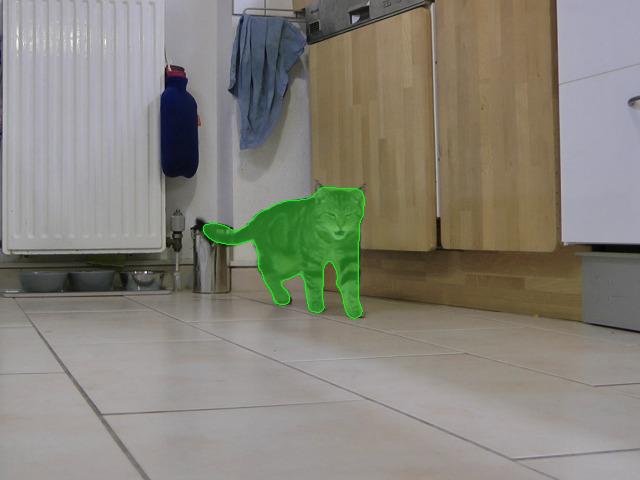}} \hfil
\subfloat{\includegraphics[width=0.187\linewidth]{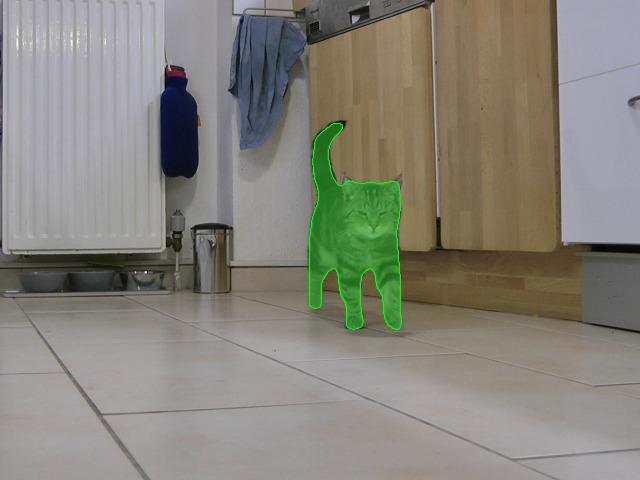}} \hfil
\subfloat{\includegraphics[width=0.187\linewidth]{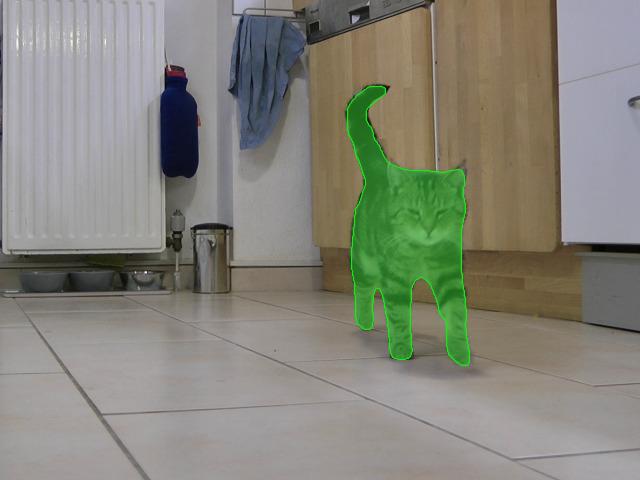}} \hfil
\subfloat{\includegraphics[width=0.187\linewidth]{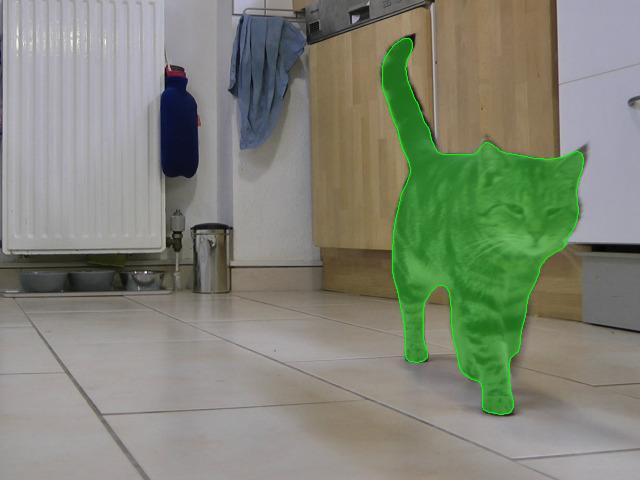}} \hfil\\
\vspace{-0.1in}
\subfloat{\includegraphics[width=0.187\linewidth]{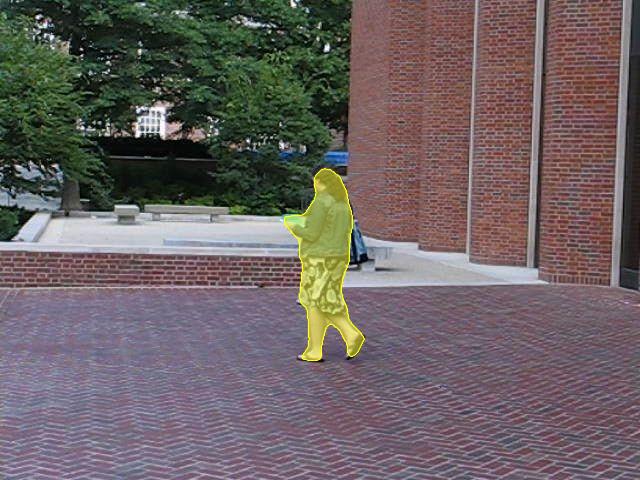}} \hfil
\subfloat{\includegraphics[width=0.187\linewidth]{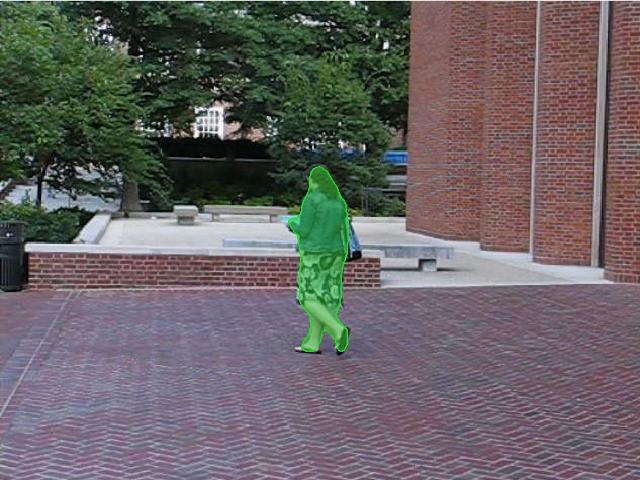}} \hfil
\subfloat{\includegraphics[width=0.187\linewidth]{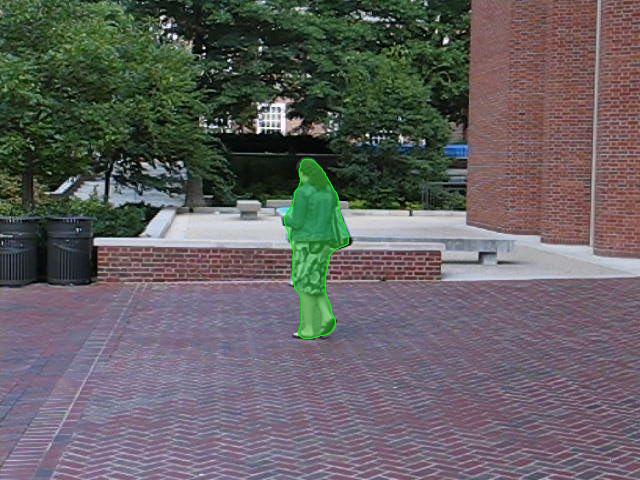}} \hfil
\subfloat{\includegraphics[width=0.187\linewidth]{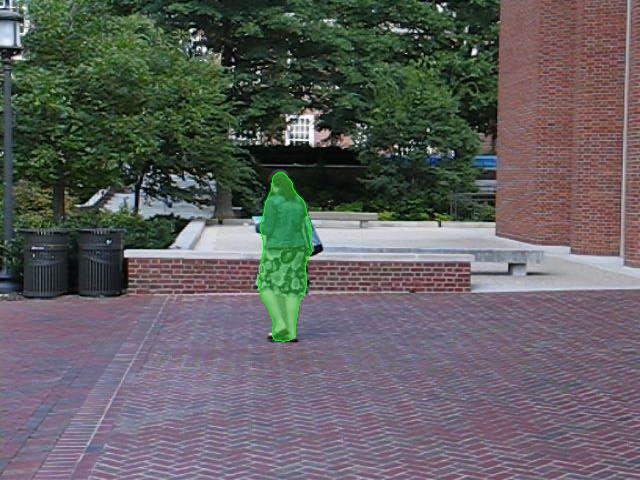}} \hfil
\subfloat{\includegraphics[width=0.187\linewidth]{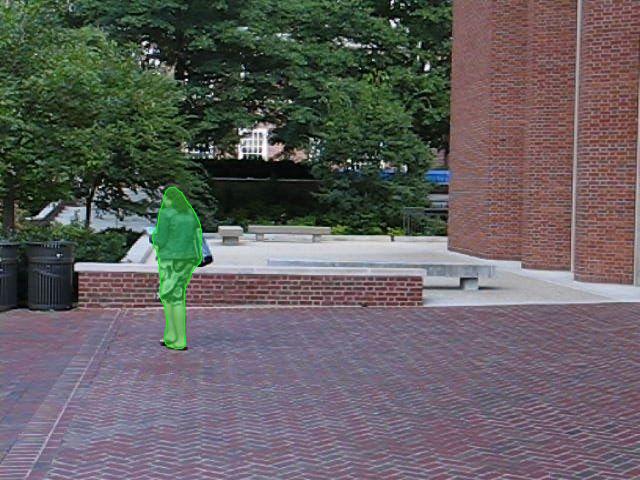}} \hfil \\
\vspace{-0.1in}
\subfloat{\includegraphics[width=0.187\linewidth]{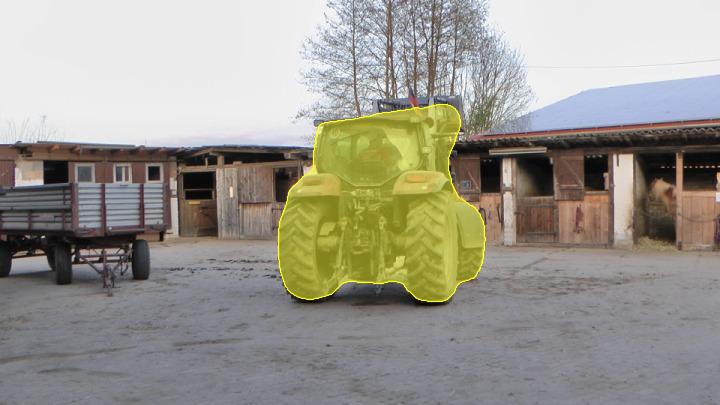}} \hfil
\subfloat{\includegraphics[width=0.187\linewidth]{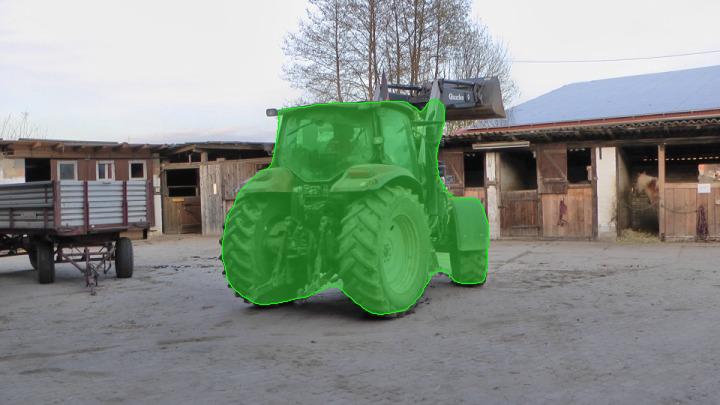}} \hfil
\subfloat{\includegraphics[width=0.187\linewidth]{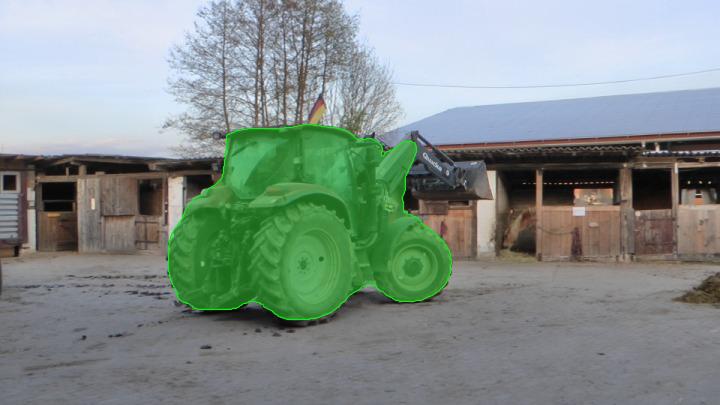}} \hfil
\subfloat{\includegraphics[width=0.187\linewidth]{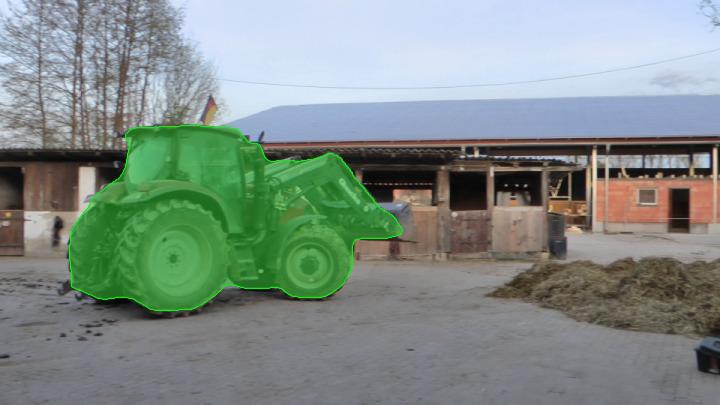}} \hfil
\subfloat{\includegraphics[width=0.187\linewidth]{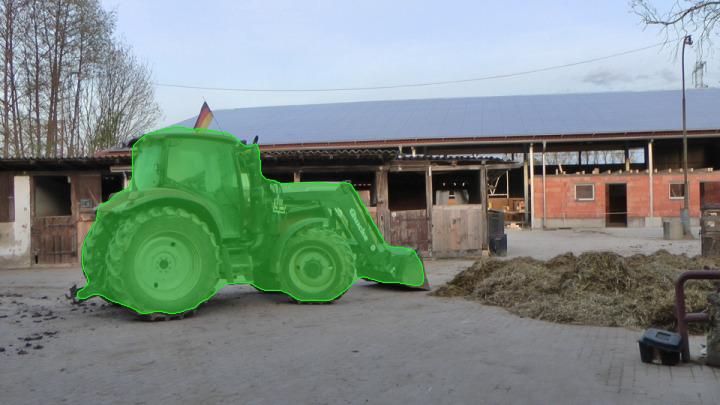}} \hfil\\
\vspace{-0.1in}
\subfloat{\includegraphics[width=0.187\linewidth]{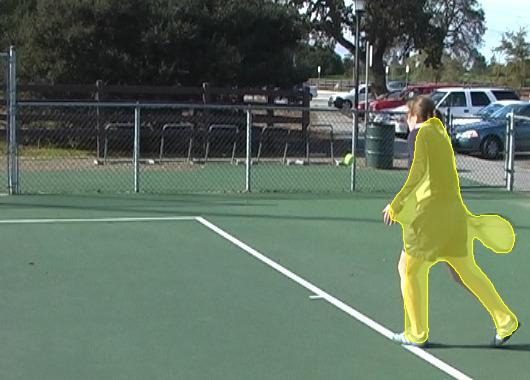}} \hfil
\subfloat{\includegraphics[width=0.187\linewidth]{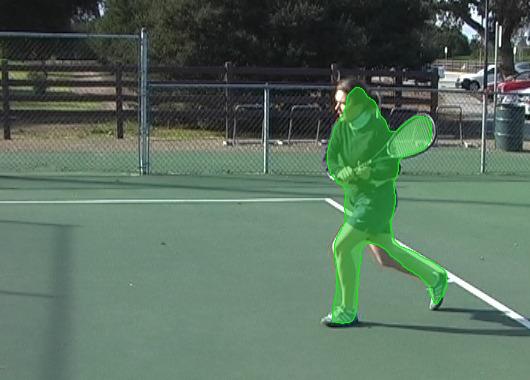}} \hfil
\subfloat{\includegraphics[width=0.187\linewidth]{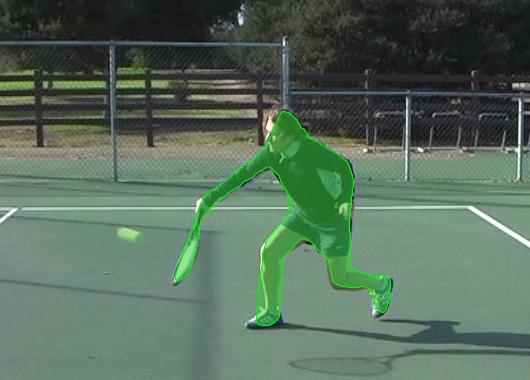}} \hfil
\subfloat{\includegraphics[width=0.187\linewidth]{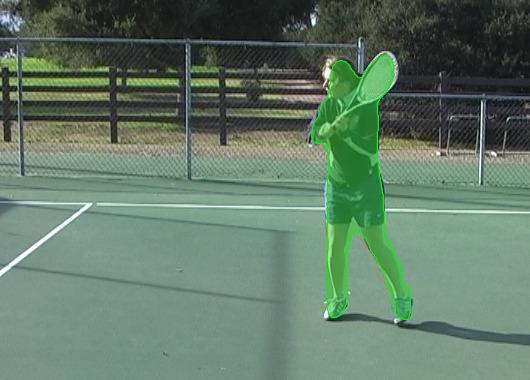}} \hfil
\subfloat{\includegraphics[width=0.187\linewidth]{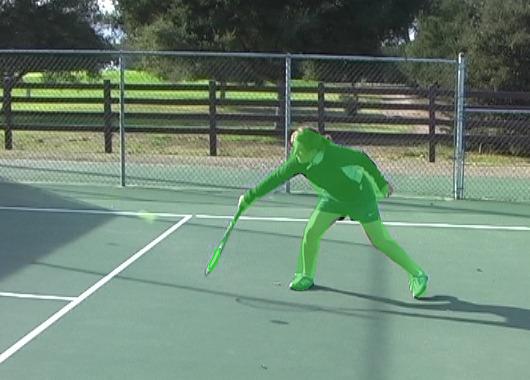}} \hfil\\
\vspace{-0.1in}
\subfloat{\includegraphics[width=0.187\linewidth]{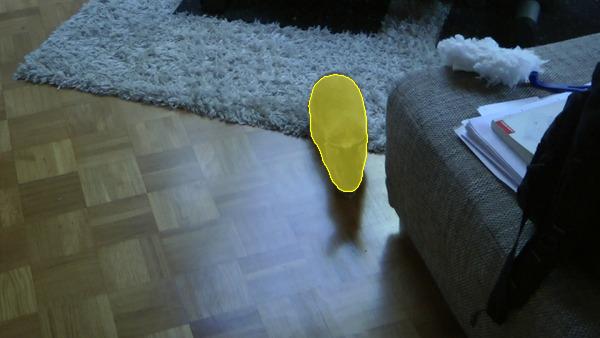}} \hfil
\subfloat{\includegraphics[width=0.187\linewidth]{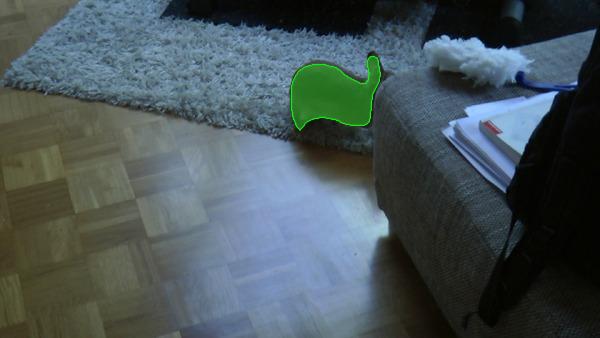}} \hfil
\subfloat{\includegraphics[width=0.187\linewidth]{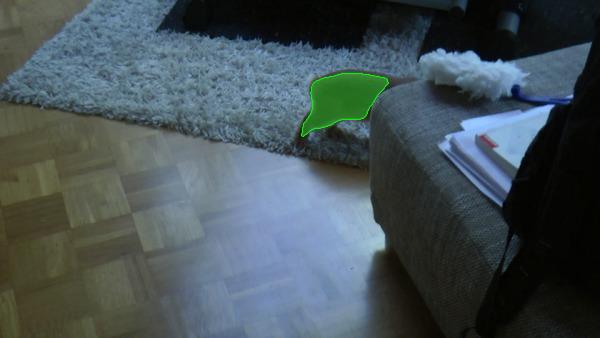}} \hfil
\subfloat{\includegraphics[width=0.187\linewidth]{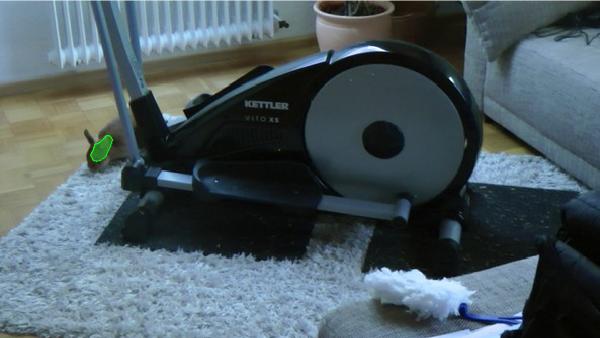}} \hfil
\subfloat{\includegraphics[width=0.187\linewidth]{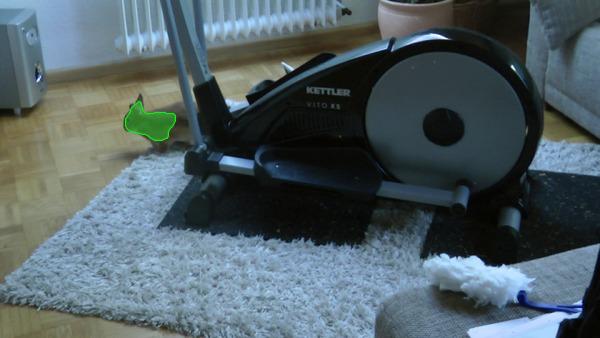}} \hfil\\
\caption{Qualitative results on FBMS dataset: The first column in yellow
is the ``pseudo ground truth'' of the first frame of each video. The
other four columns are the output segmentation masks of our proposed
approach. Best viewed in color.}\label{fig:exp}
\end{figure}

\begin{table}[t]
\centering
\caption{Comparison of the F-score and mIoU scores ($\%$) of different unsupervised
VOS approaches on the FBMS test dataset. Our method achieves
the highest compared with state-of-the-art methods} 
\label{tab:FBMS}
\begin{tabular}{|c|c|c|c|c|c|c||c|}
\hline
 & NLC \cite{faktor2014video} & FST \cite{papazoglou2013fast} & CVOS \cite{taylor2015causal} & MP-Net-V \cite{tokmakov2017learning} & LVO \cite{tokmakov2017learninglvo} & ARP \cite{koh2017primary} & Ours \\  \hline \hline
mIoU & 44.5 & 55.5 & - & - & - & 59.8 & \bf{77.9} \\ \hline
F-score & - & 69.2 & 74.9 & 77.5 & 77.8 & - & \bf{85.1} \\ \hline
\end{tabular}
\end{table}

\noindent
\subsubsection{SegTrack-v2.} 
\begin{figure}[t]
\captionsetup[subfigure]{}
\centering 
\subfloat{\includegraphics[width=0.187\linewidth]{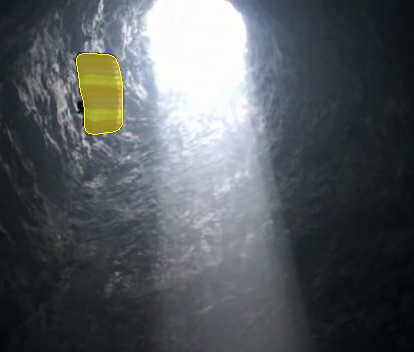}} \hfil
\subfloat{\includegraphics[width=0.187\linewidth]{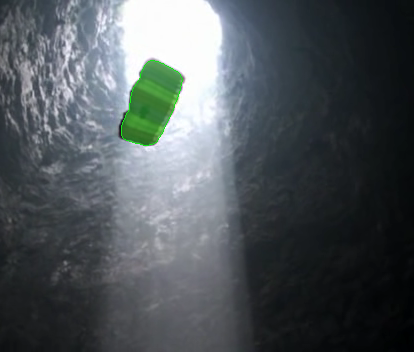}} \hfil
\subfloat{\includegraphics[width=0.187\linewidth]{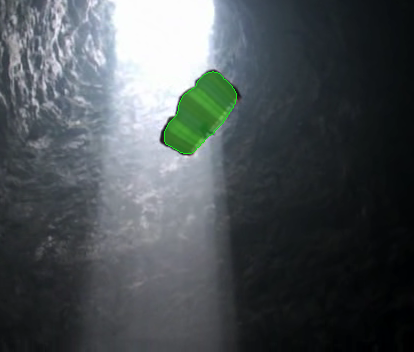}} \hfil
\subfloat{\includegraphics[width=0.187\linewidth]{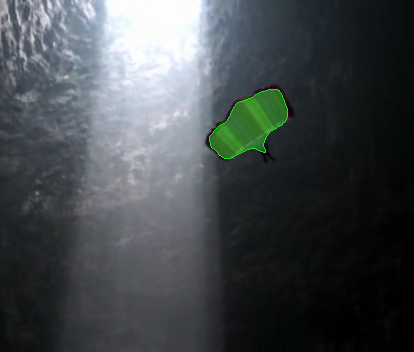}} \hfil
\subfloat{\includegraphics[width=0.187\linewidth]{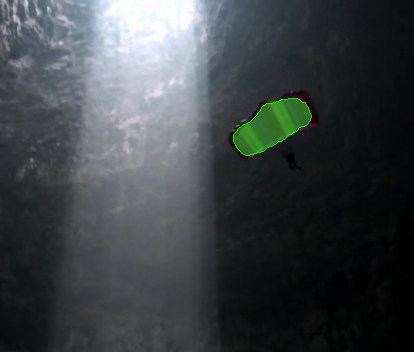}} \hfil \\
\vspace{-0.1in}
\subfloat{\includegraphics[width=0.187\linewidth]{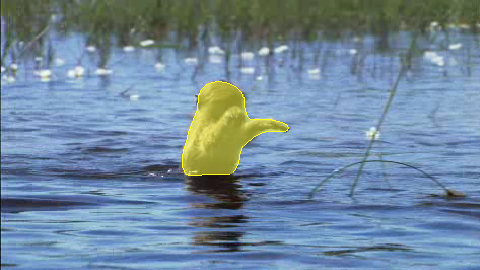}} \hfil
\subfloat{\includegraphics[width=0.187\linewidth]{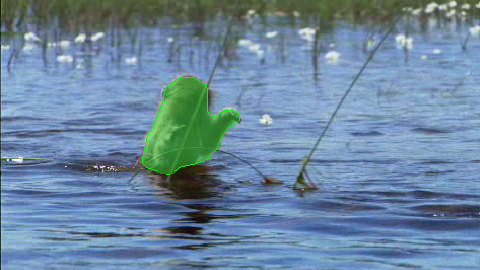}} \hfil
\subfloat{\includegraphics[width=0.187\linewidth]{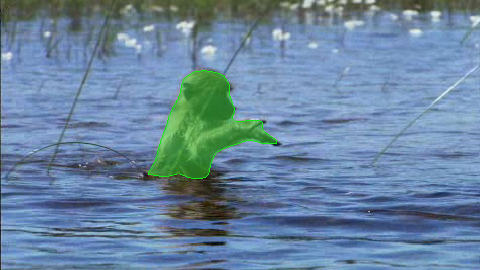}} \hfil
\subfloat{\includegraphics[width=0.187\linewidth]{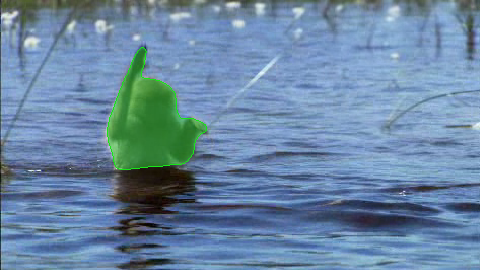}} \hfil
\subfloat{\includegraphics[width=0.187\linewidth]{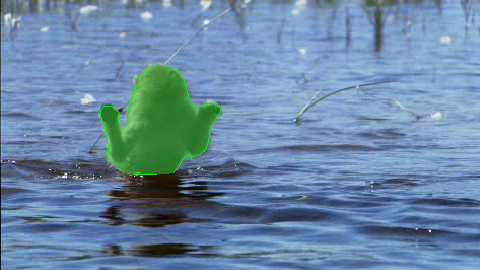}} \hfil\\
\vspace{-0.1in}
\subfloat{\includegraphics[width=0.187\linewidth]{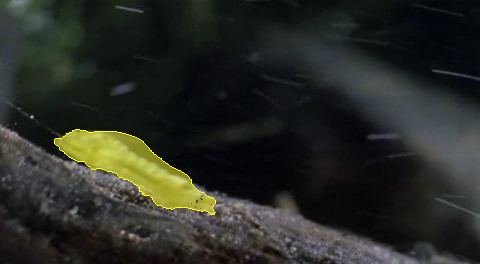}} \hfil
\subfloat{\includegraphics[width=0.187\linewidth]{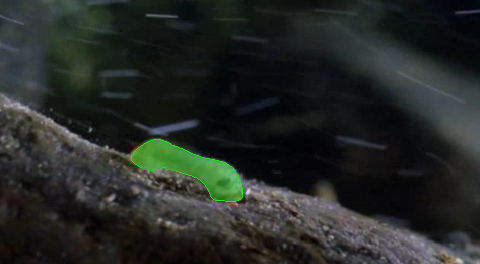}} \hfil
\subfloat{\includegraphics[width=0.187\linewidth]{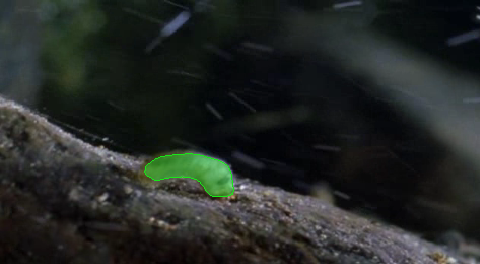}} \hfil
\subfloat{\includegraphics[width=0.187\linewidth]{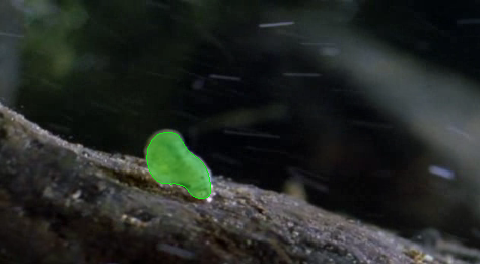}} \hfil
\subfloat{\includegraphics[width=0.187\linewidth]{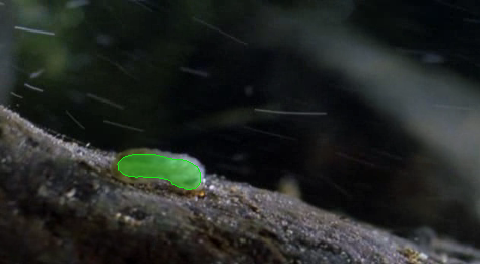}} \hfil \\
\vspace{-0.1in}
\subfloat{\includegraphics[width=0.187\linewidth]{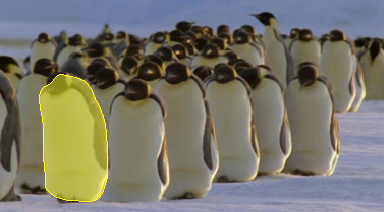}} \hfil
\subfloat{\includegraphics[width=0.187\linewidth]{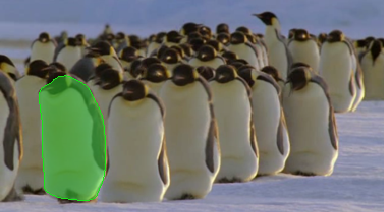}} \hfil
\subfloat{\includegraphics[width=0.187\linewidth]{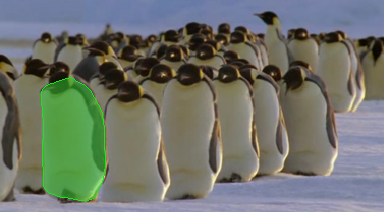}} \hfil
\subfloat{\includegraphics[width=0.187\linewidth]{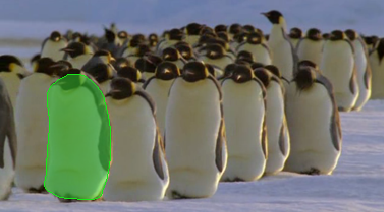}} \hfil
\subfloat{\includegraphics[width=0.187\linewidth]{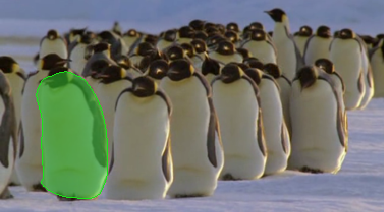}} \hfil \\
\caption{Qualitative results on SegTrack-v2 dataset: The first column in yellow
is the ``pseudo ground truth'' of the first frame of each video. The
other four columns are the output segmentation masks of our proposed
approach. Best viewed in color.}\label{fig:segtrackv2}
\end{figure}

Our method achieves mIoU of 58.7\% on this dataset,
which is higher than other methods that do well on DAVIS, CUT \cite{keuper2015motion} (47.8\%), FST \cite{papazoglou2013fast} (54.3\%), and LVO \cite{tokmakov2017learninglvo} (57.3\%). Note that we did not apply online adaptation on this dataset which could further improve the performance. Our method performs worse than NLC
\cite{faktor2014video} (67.2\%) due to low resolution of SegTrack-v2 and the fact that NLC is designed and evaluated on this dataset. We outperform NLC on both FBMS and DAVIS datasets by a large margin. Figure \ref{fig:segtrackv2} shows qualitative results of the proposed method on the SegTrack-v2. All these visual results demonstrate the effectiveness of our approach where the category of the object is not existed in MS COCO \cite{lin2014microsoft} or PASCAL VOC 2012 \cite{everingham2010pascal}. The accurate category is not needed in our approach, as long as the foreground object is consistent in the whole video. The objectness of the worm sequence in the third row cannot be detected using instance segmentation algorithm, in this case the thresholded flow magnitude is used as the pseudo ground truth mask instead.

\subsection{Ablation studies} 

Table \ref{tab:ablation} presents our ablation study on DAVIS 2016 validation set on the three major components: online adaptation, CRF \cite{chen2016deeplab} and test time data augmentation. The baseline ours-oneshot in Table \ref{tab:ablation} is the wider-ResNet trained on the PASCAL VOC 2012 dataset and the DAVIS 2016 training set. Online adaptation provides 1.4\% improvement over the baseline in terms of mIoU. Additional CRF post processing brings further 1.1\% boost in terms of mIoU. Combining with test time data augmentation (TTDA) gives the best performance of 79.3\% in mIoU which is 3.5\% higher than the baseline without any post processing. 

\begin{figure}[t]
\captionsetup[subfigure]{}
\captionsetup[subfigure]{labelformat=empty}
\centering 
\subfloat{\includegraphics[width=0.32\linewidth]{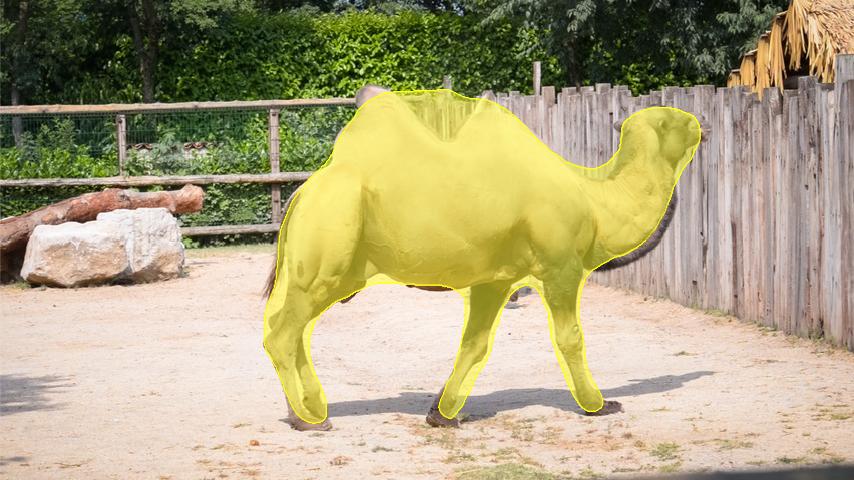}} \hfil
\subfloat{\includegraphics[width=0.32\linewidth]{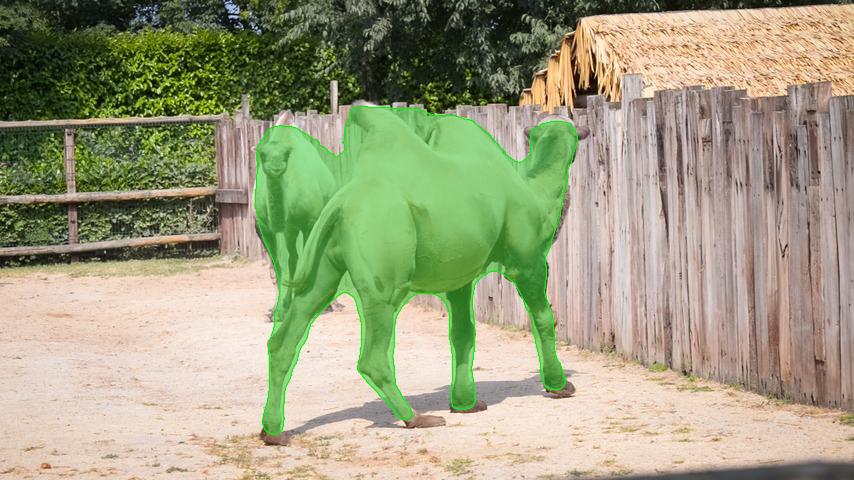}} \hfil
\subfloat{\includegraphics[width=0.32\linewidth]{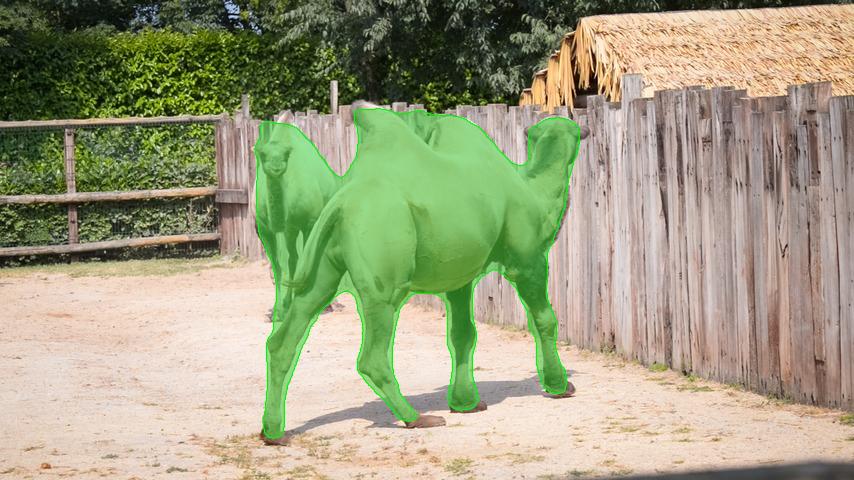}} \hfil \\
\vspace{-0.1in}
\subfloat[``pseudo ground truth'']{\includegraphics[width=0.32\linewidth]{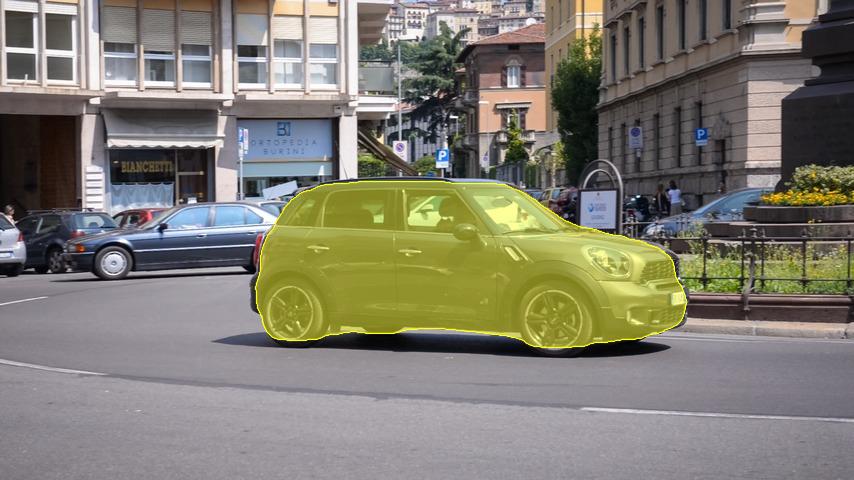}} \hfil
\subfloat[Ours-oneshot]{\includegraphics[width=0.32\linewidth]{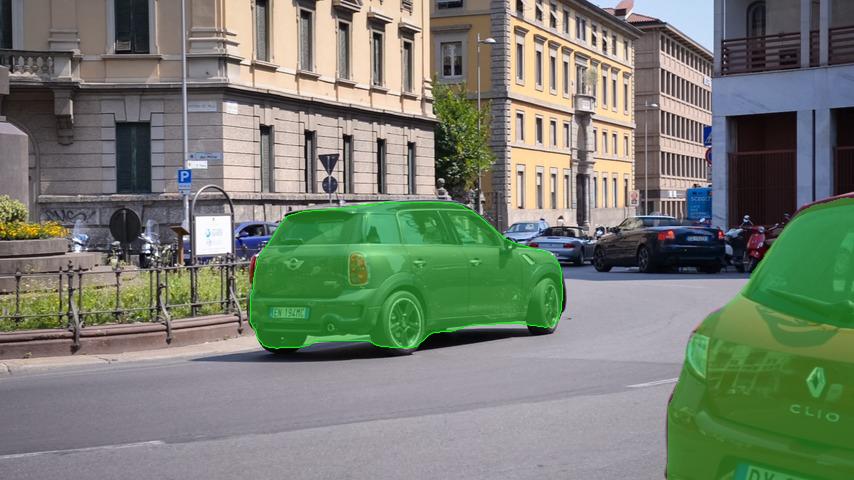}} \hfil
\subfloat[Ours-online]{\includegraphics[width=0.32\linewidth]{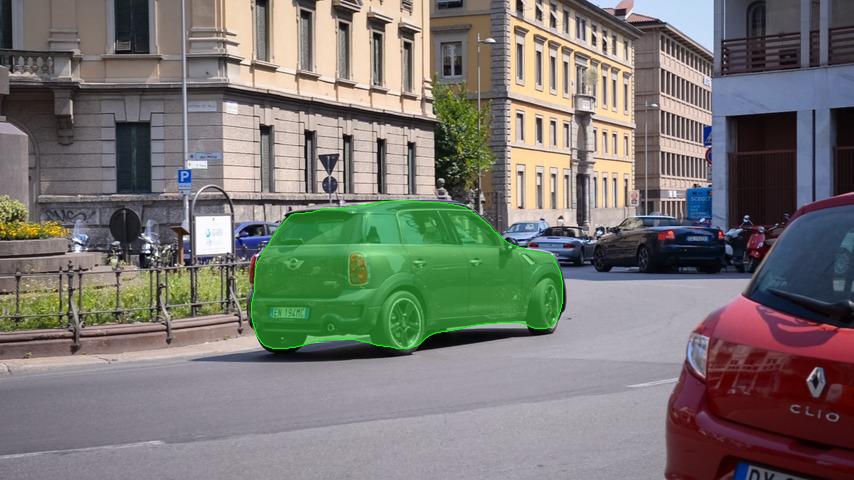}} \hfil \\
\caption{Comparison of qualitative results on two sequences, camel and car-roundabout, on DAVIS validation set: The first column
in yellow is the ``pseudo ground truth'' of the first frame of each
video. The other two columns are the output segmentation masks of our oneshot and online approaches respectively. Best viewed in color.}\label{fig:comparison}
\end{figure}

Figure \ref{fig:comparison} shows qualitative comparisons for oneshot and online approaches on the video sequences camel and car-roundabout. Our online approach outperforms our oneshot approach for the sequence car-roundabout in the second row, which is due to the right bottom pixels are considered as negative training examples from the previous frames. The additional round of fintuning is performed on the newly acquired data to remove the false positive masks. The first row shows the failure case for the two approaches, the two branches both wrongly predict the foreground mask when the moving camel is walking across the static camel. This example shows the weakness of the oneshot approaches by propagating thoughout the whole video without using motion information. 

\begin{table}[t]
\centering
\caption{Ablation study of our approach on DAVIS 2016 validation set.TTDA denotes the test time data augmentation and CRF denotes conditional random field} 
\label{tab:ablation}
\begin{tabular}{|c|c|c|c|c|}
\hline
 & Ours-oneshot & +Online & +Online +CRF & +Online +CRF +TTDA \\  \hline \hline
mIoU($\%$) & 75.8 & 77.2 & 78.3 & 79.3 \\ \hline
\end{tabular}
\end{table}

\noindent
\subsubsection{Error analysis.} To analyze the effect of the first frame tagging,
we apply OSVOS to the entire DAVIS validation set using
the pseudo ground truth and the ground truth of the first frame
respectively, the mIoUs of the entire dataset and two difficult sequences are shown in Table \ref{tab:error}. The mIoUs of
the entire DAVIS validation set is 5.5\% lower when using pseudo
ground truth of the first frame. This demonstrates that more accurate
mask prediction for the first frame can generate better segmentation
masks for the remaining frames of the whole video, which shows the potential performance
improvement when using more advanced tagging technique. 

We also erode and dilate the pseudo ground truth by 5 pixels
respectively and use the erosion and dilation masks as the new pseudo
ground truths to apply OSVOS approach to the videos. The performances
have largely degraded from 3.2\% to 10.9\% compared with those of the
original pseudo ground truth. This demonstrates accurate tagging is the
key component for our tagging and segmenting approach. 


\begin{table}[t]
\centering
\caption{Error analysis for the entire DAVIS 2016 validation set and 
two difficult videos (bmx-trees and kite-surf) }\label{tab:error}
\begin{tabular}{|c|c|c|c|c|}
\hline
Sequences & Erosion & Dilation & Pseudo Ground Truth & Ground Truth \\  \hline \hline
bmx-trees & 33.9 & 42.4 & 46.0 & 52.5  \\ 
kite-surf & 51.0 & 56.4 & 60.3 & 66.6  \\ \hline 
mIoU & 68.4 & 76.1 & 79.3 & 84.8  \\ \hline 
\end{tabular}\label{tab:error}
\end{table}

\section{Conclusion and future work}\label{sec:con}

In this paper, we present a simple yet intuitive approach for
unsupervised video object segmentation. Specifically, instead of
manually annotating the first frame like existing semi-supervised
methods, we proposed to automatically generate the approximate
annotation, pseudo ground truth, by jointly employing instance
segmentation and optical flow. Experimental results on the DAVIS, FBMS
and SegTrack-v2 demonstrate that our approach enables effective transfer
from semi-supervised VOS to unsupervised VOS and improves the mask
prediction performance by a large margin. Our error analysis shows that using better
instance segmentation has a dramatic performance boost which shows great
potential for further improvement. Our approach is able to extend from single object tracking to multiple arbitrary object tracking based on the category-agnostic ground truths or pseudo ground truths.

\bibliographystyle{splncs04}
\bibliography{egbib}
\end{document}